\documentclass[10pt,twocolumn,letterpaper]{article}

\usepackage[pagenumbers]{cvpr} % To force page numbers, e.g. for an arXiv version
\usepackage{graphicx}
\usepackage{amsmath}
\usepackage{amssymb}
\usepackage{makecell}
\usepackage{float}

\usepackage{xcolor}
\definecolor{darkblue}{rgb}{0,0.08,0.45}
\definecolor{seaborn_blue}{HTML}{0072B2}
\definecolor{seaborn_green}{HTML}{009E73}

\usepackage[pagebackref,breaklinks,colorlinks,citecolor=darkblue,linkcolor=brown,bookmarks=false]{hyperref}

\usepackage[capitalize]{cleveref}
\crefname{section}{Sec.}{Secs.}
\Crefname{section}{Section}{Sections}
\Crefname{table}{Table}{Tables}
\crefname{table}{Tab.}{Tabs.}

\def\cvprPaperID{****} % *** Enter the CVPR Paper ID here
\def\confName{CVPR}
\def\confYear{2022}

\def\paragraph#1{\textbf{\boldmath #1\,}}
\def\paranoind#1{\noindent\textbf{\boldmath #1\,}}

\usepackage{xcolor}
\definecolor{darkblue}{rgb}{0,0.08,0.45}
\definecolor{seaborn_blue}{HTML}{0072B2}
\definecolor{seaborn_green}{HTML}{009E73}

\usepackage[skip=0pt,font=small]{caption}
\usepackage{enumitem}
\makeatletter
\@namedef{ver@everyshi.sty}{}
\makeatother
\usepackage{tikz}
\usepackage{abstract}

\begin{document}

%%%%%%%%% TITLE - PLEASE UPDATE
\title{Is Self-Supervised Learning More Robust Than Supervised Learning?}

\def\inst#1{\unskip$^{#1}$}
\def\email#1{{\small\tt#1}}

\author{
Yuanyi Zhong\inst{*1},
Haoran Tang\inst{*2},
Junkun Chen\inst{1},
Jian Peng\inst{1},
Yu-Xiong Wang\inst{1}
\vspace{4pt}
\\
\inst{1} University of Illinois at Urbana-Champaign
\hspace{35pt}
\inst{2} University of Pennsylvania
\\
\email{\{yuanyiz2, junkun3, jianpeng, yxw\}@illinois.edu}
\hspace{20pt}
\email{thr99@seas.upenn.edu}
\phantom{\thanks{~Equal contribution}}
}

\twocolumn[{
\renewcommand\twocolumn[1][]{#1}
\maketitle
\vspace{-30pt}
\begin{center}
    \centering
    \captionsetup{type=figure}
    \includegraphics[width=\textwidth]{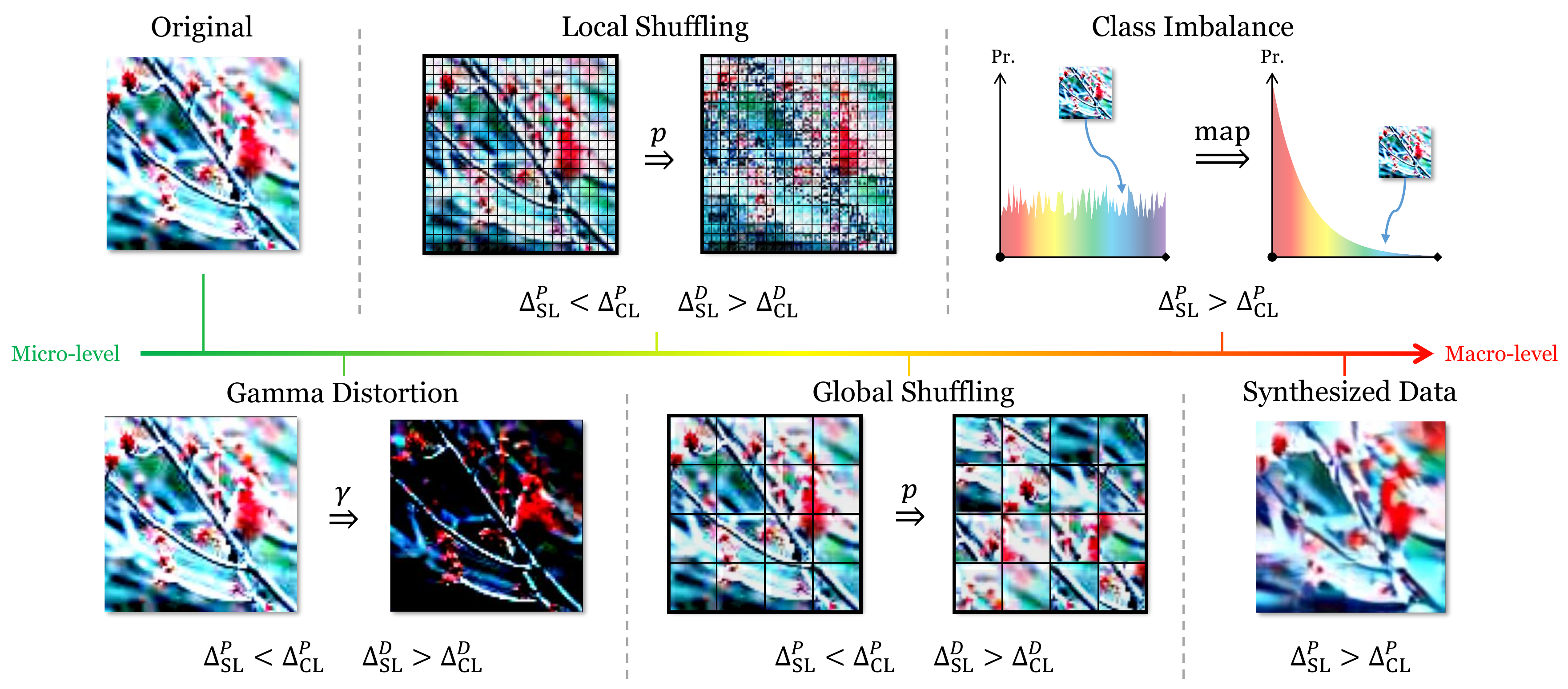}
    \captionof{figure}{We conduct a series of robustness tests based on data distribution corruptions, spanning from micro to macro level, to study the behavior of contrastive and supervised learning beyond accuracy. Our results reveal that contrastive learning is usually more robust than supervised learning to \emph{downstream} corruptions ($\Delta_{\text{\text{CL}}}^D < \Delta_{\text{\text{SL}}}^D$), while shows \emph{opposite} behaviors to \emph{pre-training dataset-level} corruptions ($\Delta_{\text{CL}}^P < \Delta_{\text{SL}}^P$) and \emph{pre-training pixel- and patch-level} corruptions ($\Delta_{\text{CL}}^P > \Delta_{\text{SL}}^P$), where $\Delta$ is the accuracy drop from uncorrupted settings.}
    \label{fig:teaser}
\end{center}
\vspace{-3pt}
}]
\saythanks

%%%%%%%%% ABSTRACT
\begin{abstract}
Self-supervised contrastive learning is a powerful tool to learn visual representation without labels. Prior work has primarily focused on evaluating the recognition accuracy of various pre-training algorithms, but has overlooked other {\em behavioral} aspects. In addition to accuracy, {\em distributional robustness} plays a critical role in the reliability of machine learning models. We design and conduct a series of robustness tests to quantify the behavioral differences between contrastive learning and supervised learning to downstream or pre-training data distribution changes. These tests leverage data corruptions at multiple levels, ranging from pixel-level gamma distortion to patch-level shuffling and to dataset-level distribution shift. Our tests unveil intriguing robustness behaviors of contrastive and supervised learning. On the one hand, under downstream corruptions, we generally observe that contrastive learning is surprisingly more robust than supervised learning. On the other hand, under pre-training corruptions, we find contrastive learning vulnerable to patch shuffling and pixel intensity change, yet less sensitive to dataset-level distribution change. We attempt to explain these results through the role of data augmentation and feature space properties. Our insight has implications in improving the downstream robustness of supervised learning.
% \keywords{Contrastive Learning, Robustness, Representation Learning}
\end{abstract}

%%%%%%%%% BODY TEXT
\section{Introduction}
\label{sec:intro}

In recent years, self-supervised contrastive learning (CL) has demonstrated tremendous potential in learning generalizable representations from unlabeled datasets \cite{chen2020simple,he2020momentum,grill2020bootstrap,chen2021exploring,caron2020unsupervised,zhong2021pixel}. Current state-of-the-art CL algorithms can learn representations from ImageNet \cite{deng2009imagenet} that match or even exceed the accuracy of their supervised learning (SL) counterparts in terms of ImageNet linear evaluation and downstream task performance \cite{chen2020simple,he2020momentum,grill2020bootstrap,chen2021exploring,caron2020unsupervised}. 

However, beyond accuracy, relatively less attention is paid on comparing other {\em behavioral differences} between contrastive learning and supervised learning. Robustness is an important aspect to evaluate machine learning algorithms. For example, robustness to long-tail or noisy training data allows the learning algorithm to work well in a wide variety of imperfect real-world scenarios \cite{wang2017learning}. Robustness of the model output across training iterations enables anytime early-stop \cite{hu2019learning} and smoother continual learning \cite{shen2020towards}. Robustness to input corruptions at test-time plays an important role in reliable deployment of trained models in safety-critical applications, as signified by the existence of adversarial examples \cite{goodfellow2014explaining,salman2020adversarially} and the negative impact of domain shift \cite{zhao2019learning}.

In this paper, we investigate whether CL and SL behave robustly to data distribution changes. In particular, how does the change in \emph{data} affect the behavior of algorithms? Do SL and CL behave similarly? 
To this end, we design a wide-spectrum of corruptions as shown in Figure~\ref{fig:teaser} to alter data distribution and conduct comprehensive experiments, with different backbones (ResNet \cite{he2016deep}, ViT\cite{dosovitskiy2020image,touvron2021training}), CL algorithms\footnote{While CL algorithms are the main focus here, the generative SSL approach MAE \cite{he2021masked} is also investigated in the downstream robustness test.} (SimCLR \cite{chen2020big}, MoCo \cite{chen2020improved,chen2021empirical}, SimSiam \cite{chen2021exploring}, BYOL\cite{grill2020bootstrap}, BarlowTwins \cite{zbontar2021barlow}, DeepCluster-v2, SwAV \cite{caron2020unsupervised}, DINO \cite{caron2021emerging}), and datasets (CIFAR \cite{krizhevsky2009learning}, ImageNet \cite{deng2009imagenet}, STL-10 \cite{coates2011analysis}, Cars \cite{krause20133d}, Aircrafts \cite{maji2013fine}). The corruptions are carefully selected to be multi-level, corrupt different structural information (not necessarily human-recognizable), and are rooted in prior literature: pixel-level gamma distorts intensity distribution, patch-level shuffle corrupts spatial structure \cite{ge2021robust,neyshabur2020being,zhang2016understanding}, and dataset-level class imbalance \cite{liu2021self,liu2019large,Samuel2021DistributionalRL} and GAN (generative adversarial network) synthesis \cite{jahanian2021generative} shift the overall distribution.

Our main results consist of two sets of experiments: The first set investigates the \emph{downstream} robustness of pre-trained models towards corruptions of downstream data. The second set studies the robustness under \emph{pre-training} data corruptions.
When the accuracy degradation of an algorithm to some corruption is large, it suggests that the algorithm may leverage such information as learning signal.
Note that our work is inspired by \cite{zhang2016understanding,ribeiro2020beyond} and follows a similar \emph{empirical exploratory analysis}, rather than a regular (adversarial) robustness paradigm.

We deliver a set of \textbf{intriguing new discoveries}.
We generally observe that CL is consistently more robust than SL to \emph{downstream} corruptions. On the other hand, contrastive learning on corrupted \emph{pre-training} data leads to \emph{diverging} observations depending on the corruption type: CL is more robust to dataset-level corruption than SL, but less so to pixel- and patch-level corruptions.

While the exact reason why pre-trained CL models are more robust to downstream corruptions remains unclear, our analysis of the learning dynamics through feature space metrics reveals one piece of the puzzle: CL yields larger \emph{overall} and steadily-increasing \emph{per-class feature uniformity} and higher stability than SL. 
The instance-level CL objective might capture richer sets of features that are not limited to semantic classes. Therefore, the per-class uniformity or intra-class variation is not compressed as hard as in SL. Such hypothesis aligns well with several recent attempts to understand CL better \cite{zhao2020makes,chen2020intriguing,liu2021self}. 
An immediate consequence of our insight is an improvement to supervised pre-training by adding a uniformity regularization term to explicitly promote intra-class variance. As a proof-of-concept, we are able to improve the test-time data corruption robustness of a ResNet-18 \cite{he2016deep} model pre-trained on CIFAR-10 \cite{krizhevsky2009learning} using the joint objective.
As for CL's vulnerability to pre-training data corruption types such as patch shuffling, we hypothesize that the interference with random crop augmentation is the main culprit. To verify our intuition, we switch the order of data corruption and standard augmentations, and find that CL recovers a substantial amount of robustness.

We summarize {\bf our contributions} as follows. (1) We design extensive robustness tests to study the behavioral differences of CL and SL systematically.
(2) We discover diverging robustness behaviors between CL and SL, and even among different CL algorithms.
(3) We offer initial analyses and explanations for such observations, and show a simple way to improve the downstream robustness of supervised learning by encouraging uniformity.
We claim our paper as an empirical study. We hope our findings can serve as an initial step to fully understand CL's behaviors beyond accuracy and inspire more future studies to explore such aspects through larger-scale experiments and theoretical analysis.

\section{Related Work}

\paranoind{Supervised Learning (SL).}
Supervised deep learning, large labeled dataset, and computation have been a success recipe for cracking many visual recognition problems \cite{russakovsky2015imagenet,he2016deep}. Typically, one first collects a large-scale dataset for the target vision problem and crowdsources labels for the specific task from human annotators. A machine learning model, \eg, a neural net, is then trained by minimizing a loss function defined on the prediction-label pairs.

\smallskip\paranoind{Self-Supervised Learning (SSL) and Contrastive Learning (CL).}
Remarkable progress has been made in self-supervised representation learning from unlabeled datasets
\cite{chen2020simple,he2020momentum,grill2020bootstrap,chen2021exploring,caron2020unsupervised}. This paper focuses on a particular kind of SSL algorithm, contrastive learning, that learns augmentation invariance with a Siamese network. To prevent trivial solution, contrastive learning pushes negative examples apart (MoCo \cite{he2020momentum,chen2020improved,chen2021empirical}, SimCLR \cite{chen2020simple,chen2020big}), makes use of stop-gradient operation or asymmetric predictor without using negatives (SimSiam \cite{chen2021exploring}, BYOL \cite{grill2020bootstrap}, DINO \cite{caron2021emerging}), or leverages redundancy reduction (BarlowTwins \cite{zbontar2021barlow}) and clustering (DeepCluster-v2 and SwAV \cite{caron2020unsupervised}). In addition to augmentation invariance, generative pre-training \cite{ramesh2021zero,bao2021beit,he2021masked} and visual-language pre-training \cite{radford2021learning} are promising ways to learn transferable representations.

There is a growing body of literature on understanding self-supervised learning. \cite{wang2021understanding} decomposes the contrastive objective into alignment (between augmentations) and uniformity (across entire feature space) terms. Uniformity can be thought of as an estimate of the feature entropy, which we leverage as a metric to study the feature space dynamics during training. \cite{wang2020understanding} makes connection between uniformity and the temperature parameter in a contrastive loss, and finds that a good temperature can balance uniformity and tolerance of semantically similar examples. \cite{zhao2020makes} discovers that SSL transferring better than SL can be due to better low- and mid-level features, and the intra-class invariance objective in SL weakens transferability by causing more pre-training and downstream task misalignment. \cite{ericsson2021well} studies the downstream task accuracy of a variety of pre-trained models and finds that SSL outperforms SL on many tasks. \cite{cole2021does} investigates the impact of pre-training data size, domain quality, and task granularity on downstream performance. \cite{chen2020intriguing} identifies three intriguing properties of contrastive learning: a generalized version of the loss, learning with the presence of multiple objects, and feature suppression induced by competing augmentations. Our work falls into the same line of research that attempts to understand SSL better. However, we investigate from the angle of \emph{robustness behavior comparison} between SSL/CL and SL.

\paranoind{Robustness and Data Corruption.}
The success of learning algorithms is often measured by some form of task accuracy, such as the top-1 accuracy for image classification \cite{deng2009imagenet,krizhevsky2009learning,coates2011analysis,wah2011caltech}, or the mean average precision for object detection \cite{he2020momentum,zhong2021dap,zhong2020boosting}. Beyond accuracy, robustness is another important measure \cite{hendrycks2018benchmarking}, and there are benchmarks and metrics proposed for SL \cite{nado2021uncertainty}. Robustness is more and more studied in SSL settings \cite{chuang2022robust,goyal2022vision}. \cite{chuang2022robust} tries to improve CL's robustness to noisy positive views. Recent large-scale study reveals that vision models are more robust and fair when pre-trained on uncurated images without supervision \cite{goyal2022vision}. We use ``robustness'' to refer to the ability of learning algorithms to cope with systematic train or test data corruptions. Under the supervised setting, deep models are shown to train successfully (albeit not to generalize) under pixel shuffling corruption and random labels, even though they are not human-recognizable anymore \cite{zhang2016understanding}.

Adversarial robustness \cite{szegedy2013intriguing,goodfellow2014explaining,madry2018towards,shafahi2019adversarial,chen2020adversarial} is a related but different concept, which refers to the model's ability to defend against adversarial attacks. An adversarial attack \cite{szegedy2013intriguing,goodfellow2014explaining} is a perceptually indistinguishable perturbation to a \emph{single} image that fools the model. Adversarial training \cite{madry2018towards,shafahi2019adversarial} is a technique to achieve adversarial robustness. Self-supervised perturbation is explored in adversarial attack and training \cite{naseer2020self,kim2020adversarial}. \cite{hendrycks2019using} shows that SSL models possess better adversarial robustness. \cite{fan2021does} improves the adversarial robustness transferability of CL.
Our definition of robustness differs from adversarial robustness -- we use robustness to analyze the tolerance of learning methods to \emph{systematic data corruptions} (rather than per-image imperceptible perturbation).

There are many types of data corruptions in prior work. The most common data corruptions, such as random resizing and cropping, flipping, and color jittering, appear as data augmentation in SL and SSL \cite{he2016deep,he2020momentum,chen2020simple}. The learned representation is encouraged to be invariant to such corruptions. \cite{hendrycks2018benchmarking} proposes a set of corruptions complementary to ours.
Block shuffling (our image global shuffling) has been used to study what is transferred in transfer learning \cite{neyshabur2020being} and as negative views with diminished semantics in contrastive learning \cite{ge2021robust}.
\cite{cole2021does} tampers data quality in SimCLR and SL training by salt-and-pepper noise, JPEG, resizing, and downsampling, and tests on clean data. We use a different set of data corruptions and test on the corrupted data as well. A recent work \cite{jahanian2021generative} also studies generative models as an alternative data source for contrastive learning. They focus on comparison with real data, while we emphasize the behavior difference of SSL and SL in response to the generative data source. Feature backward-compatibility \cite{shen2020towards} is related to our stability analysis of feature dynamics. Recently, \cite{goyal2021self} studies the effectiveness of SSL on uncurated class-imbalanced data. \cite{liu2021self} also notices that SSL tends to be more robust to class imbalance than SL.
We bring extra insights over them. We consider \emph{both pre-training and downstream} robustness and compare \emph{CL and SL behaviors}, while \cite{goyal2021self} only focuses on downstream and compares dataset scale. Our investigation suggests that pre-train behavior can be \emph{opposite} to downstream. \cite{liu2021self} only studies class imbalance, but we consider broader corruptions.

\section{Method}

We evaluate the robustness of different pre-training algorithms by observing the impact of a series of carefully-designed data corruptions. To what extent will such corruptions influence the performance? Will there be consistent trends that depend on the type of the corruptions? And will there be a behavioral difference between CL and SL?

\subsection{Robustness Tests}

The common way of using CL or SL trained models is through the pre-training and fine-tuning paradigm \cite{chen2020simple,he2020momentum,zhong2021dap,zhong2021pixel}. A neural net backbone is pre-trained on a large-scale dataset such as ImageNet \cite{deng2009imagenet}, and transferred to initialize downstream models. Therefore, it is natural to consider the impact of data corruptions in both the pre-training phase and the fine-tuning phase. Specifically, we perform the following two complementary types of tests.

\smallskip\paranoind{Robustness Test I: Downstream data corruption.}
In this test, the pre-training algorithm is run on the clean version of the pre-training dataset. For a given downstream dataset, we evaluate the pre-trained model's accuracy on its original version and various corrupted versions. This assesses the robustness of the algorithm by looking at the pre-trained model's robustness behaviors.

\smallskip\paranoind{Robustness Test II: Pre-training data corruption.}
This assesses the algorithm's robustness to pre-training data corruptions. We run the pre-training algorithm on the corrupted version of the dataset, and then evaluate the final model's accuracy on either the corrupted test set or the original test set. The test set can be in-domain (the same domain as the train set) or out-domain (a different domain from the train set). Since data corruption destroys certain information by design, the model pre-trained on the corrupted data is expected to yield degraded performance than the model pre-trained on the original dataset.

\smallskip\paranoind{Robustness Metric.}
In both cases, the robustness is measured by the degradation in accuracy caused by certain data corruption. An algorithm is more robust if the degradation is smaller. Denote $\mathcal{D}_{\text{original}}$ as the original dataset and $\mathcal{D}_{\text{corrupted}}$ as the corrupted dataset. We use the same notation for both pre-training and downstream corruptions. For an algorithm $\text{Alg} \in \{\text{CL}, \text{SL}\}$, we define
\begin{equation}
    \Delta(\text{Alg}) = \frac{
    \text{Acc}(\text{Alg}, \mathcal{D}_\text{original}) - \text{Acc}(\text{Alg}, \mathcal{D}_\text{corrupted})
    }{\text{Acc}(\text{Alg}, \mathcal{D}_\text{original})}
    .
\end{equation}

We use two methods to obtain the test accuracy in the above equation. The first is \textbf{linear evaluation}, where we train a linear classifier on top of the representation learned by the pre-training algorithm on the train split, and then evaluate the classifier on the test split. The second is \textbf{KNN evaluation}. We use the weighted KNN classifier from \cite{wu2018unsupervised}, where the prediction is the exponential-distance weighted average of the $K$ nearest neighbors in the train split of any test data point, measured by the normalized feature vectors. The KNN evaluation effectively leverages an non-parametric classifier, therefore no classifier training is required. Depending on the context, we use either linear or KNN evaluation in our experiments.
The essential question we would like to ask is whether $\Delta(\text{CL})$ is consistently larger or smaller than $\Delta(\text{SL})$ across different data corruptions.

\subsection{Types of Data Corruption}
There is a natural hierarchy of data corruptions ranging conceptually from {\em micro-level} to {\em macro-level}. We describe our choices below, which are also illustrated in Figure~\ref{fig:teaser}. Note that our data corruption is different from data augmentation. In data augmentation, the corruption is applied randomly on a per-image basis. In our case, a fixed random transformation (\eg, the gamma in gamma distortion or the permutation order in shuffling) is decided first and then applied consistently across all images. We effectively transform the entire dataset with the corruption method.

We emphasize that our purpose is not to study human-recognizable distortions, but to evaluate pre-training algorithms' behavior under distortions. To this end, our collection of corruptions is designed to be representative and comprehensive: while some of them are practical (gamma, imbalance), others are purposefully introduced to distort certain structural information (shuffle). The similar flavor of behavior study was seen in \cite{zhang2016understanding}.

\smallskip
\paranoind{Pixel-Level Corruption.}
The pixel intensity distribution is altered, but neither the spatial layout of each image nor the overall data distribution is changed. The popular color jittering augmentation randomly modifies the brightness, contrast, saturation, and hue of training images. Here, we deliberately pick gamma distortion, as it is not already part of the conventional data augmentation pipeline.

\begin{itemize}[topsep=0pt,itemsep=0pt,leftmargin=0.9em]
    \item 
\textbf{Gamma distortion:} Gamma distortion remaps each R,G,B pixel intensity ($\in [0,255]$) according to $ x \to \lfloor 255\times (x/255) \rfloor^\gamma $, where $\gamma > 0$ is a tunable parameter. $\gamma=1$ recovers the original intensity. Larger or smaller $\gamma$ shifts the intensities darker or brighter, respectively. Due to quantization error, there will be part of the intensity information lost during the process.
\end{itemize}

% \smallskip
\paranoind{Patch-Level Corruption.}
Pixel-level corruption does not change the spatial layout of pixels. The next type of corruption in the hierarchy is at patch level. Inspired by prior work \cite{zhang2016understanding}, we employ global and local shuffling. Note that pixel shuffling is not commonly used in the standard augmentation pipeline of visual recognition training \cite{he2016deep}. We are curious what kind of behaviors CL and SL will exhibit under the patch-level shuffling data corruption.

\begin{figure}[t]
  \centering
  \includegraphics[width=0.75\linewidth]{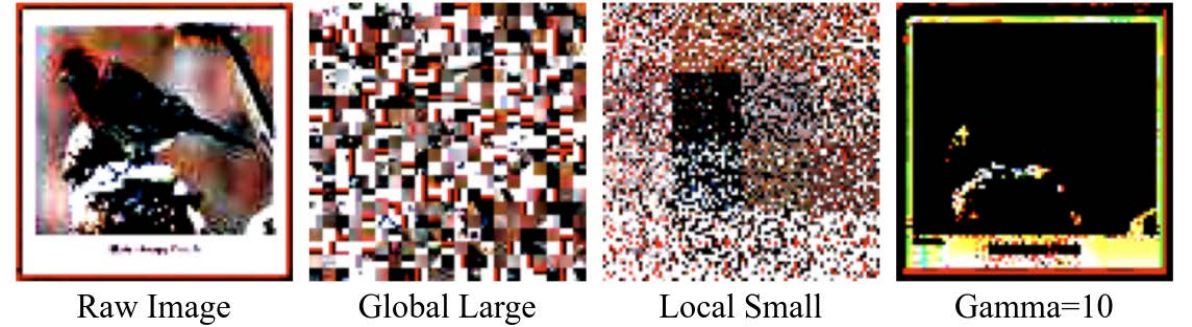}
  \caption{Examples of extreme data corruptions. ``Small'' and ``large'' refer to the number of patches we divide an image into, which varies with different datasets.}
  \label{fig:extreme}
  \vspace{-10pt}
\end{figure}

\begin{itemize}[topsep=0pt,itemsep=0pt,leftmargin=0.9em]
    \item 
\textbf{Global shuffling:} Global shuffling breaks down the image into patches and shuffles the patches according to a fixed random order. Specifically, suppose the image size is $s\times s$ and the size of each patch is $p\times p$, and then the image is divided into $s/p \times s/p$ patches. Global shuffling destroys the global spatial structure of an image but preserves the local structure. The image becomes less structured with a smaller patch size.
    \item
\textbf{Local shuffling:} Inversely to global shuffling, local shuffling randomly permutes the pixels inside each local $p\times p$ patch by a fixed random order, but keeps the global ordering of patches. It damages the local image structure while preserving the overall global structure. The image becomes less structured with a larger patch size.
\end{itemize}

% \smallskip
\paranoind{Dataset-Level Corruption.}
Finally, we consider corruptions happening at the whole dataset distribution level. The previous two types of corruptions change the images, but not the distribution of the images.

\begin{itemize}[topsep=0pt,itemsep=0pt,leftmargin=0.9em]
    \item 
\textbf{Synthesized data:} Synthesized data is popularizing (\eg, DALL$\cdot$E 2 \cite{ramesh2022hierarchical}) and studied to replace real data \cite{jahanian2021generative}. We utilize generative methods such as a conditional GAN \cite{karras2020training} to generate a synthesized dataset $\mathcal{D}_{\text{GAN}}$ and replace $\mathcal{D}_{\text{original}}$. We can then measure and compare $\Delta(\text{Alg})$ between these two datasets of different distributions. Oftentimes, the generated distribution is not perfectly aligned with the real distribution, therefore training with the generative data source may lead to degradation in accuracy of clean data or downstream performance.
    \item
\textbf{Class imbalance:} Real-world data often follows a long-tail distribution of semantic classes, where a few common classes have lots of examples while many tail classes have few examples~\cite{kang2019decoupling,Samuel2021DistributionalRL}. However, benchmark datasets such as CIFAR and ImageNet are curated and class-balanced. We consider the widely-used variant of ImageNet, ImageNet-LT (long-tail) \cite{liu2019large}, with maximally 1280 images and minimally 5 images per class. As a comparison, we construct ImageNet-UF (uniform), a class-balanced subset of ImageNet which contains the same number of images as ImageNet-LT (115K). Then we test whether moving from pre-training on ImageNet-UF to ImageNet-LT would lead to different behaviors between CL and SL.
\end{itemize}

\subsection{Experiment Setup}

\paranoind{Datasets.} We pre-train on CIFAR-10\cite{krizhevsky2009learning} and ImageNet\cite{deng2009imagenet} variants to evaluate the robustness differences between CL and SL methods subject to pre-training data corruptions. We use CIFAR-10/100 \cite{krizhevsky2009learning}, STL-10 \cite{coates2011analysis}, and two fine-grained classification datasets, Cars\cite{krause20133d} and Aircrafts\cite{maji2013fine}, to analyze the performance of the pre-trained models on the corrupted downstream tasks. For fair comparisons, we use the same data augmentation across methods when we need to train any model.

\smallskip\paranoind{Models and Algorithms.}
We benchmark a variety of self-supervised contrastive learning algorithms. These methods are carefully sampled to be representative. They include contrastive learning with negatives (SimCLR-v2\cite{chen2020simple,chen2020big}, MoCo-v2\cite{he2020momentum,chen2020improved}), without negatives (SimSiam\cite{chen2021exploring}, and the momentum counterpart, BYOL\cite{grill2020bootstrap}), with redundancy reduction (BarlowTwins\cite{zbontar2021barlow}), and contrasting clustering assignments (DeepCluster-v2\cite{caron2020unsupervised}, SwAV\cite{caron2020unsupervised}). We test both CNN (standard ResNet-18/50\cite{he2016deep}) and Vision Transformer (ViT)\cite{dosovitskiy2020image} backbones. For transformers, we leverage pre-trained models on ImageNet \cite{deng2009imagenet} from ViT\cite{dosovitskiy2020image}, DeiT\cite{touvron2021training}, DINO\cite{caron2021emerging}, MoCo-v3\cite{chen2021empirical}, and MAE\cite{he2021masked} (which makes an interesting comparison as it is based on reconstruction rather than contrasting).

\def\Dn#1{{#1\%}}
\def\MinNumber{30.0}
\def\MaxNumber{60.0}
\def\Dc#1{%
 \pgfmathsetmacro{\PercentColor}{max(min(100.0*(#1 - \MinNumber)/(\MaxNumber-\MinNumber),100.0),0.00)} %
 \colorbox{yellow!\PercentColor}{#1\%}
}

\begin{table}[t]
\caption{Robustness Test I: \emph{downstream} \emph{pixel- and patch-level} corruptions with \emph{ResNet-50} backbone \cite{he2016deep}. The models are downloaded from corresponding official websites and pre-trained on uncorrupted ImageNet (`IN Acc:' reference ImageNet Val accuracy; Suffix `-a/b:' two models of the same algorithm). We consider 5 downstream datasets. For each dataset, we report the averages of 6 corruption settings: gamma distortion $\gamma=\{0.2,5\}$, patch global and local shuffle ($p=\{4,\text{image\_size}/4\}$). The image size is 32 for C-10/100, 96 for STL-10, and 256 for the rest; the corrupted images are resized to 224 as input to the network. We compute the KNN accuracy (K=50 for C-10/100 and STL-10, K=5 for others) on corrupted test sets and report the relative drops $\Delta$ to the uncorrupted versions. Avg $\Delta$ is the average over the 5 datasets (darker shades indicate higher drops).
This table only shows $\Delta$. Please refer to Appendix \ref{sec:b.3} for the accuracy numbers.
Contrastive learning models generally show lower accuracy drops and therefore higher downstream robustness than supervised models.}
\label{tab:downstream_resnet}
\setlength{\tabcolsep}{5.2pt}
\centering
\footnotesize
\setlength{\tabcolsep}{3pt}
\scalebox{0.9}{%
\begin{tabular}{@{}lllllllc@{}}
\hline
Pre-train Alg & IN Acc & C-10 & C-100 & STL-10 & Car-196 & Air-70 & Avg $\Delta\downarrow$ \\
\hline
Sup-a & 76.1 & \Dn{31.5} & \Dn{45.3} & \Dn{31.0} & \Dn{51.2} & \Dn{39.9}   & \Dc{39.8} \\
Sup-b & 75.5 & \Dn{32.1} & \Dn{47.2} & \Dn{31.9} & \Dn{53.2} & \Dn{39.2}   & \Dc{40.7} \\
\hline
BYOL & 72.3 & \Dn{29.3} & \Dn{43.0} & \Dn{29.0} & \Dn{42.9} & \Dn{33.8} & \Dc{35.6} \\
SimSiam & 68.3 & \Dn{27.8} & \Dn{40.8} & \Dn{29.3} & \Dn{41.5} & \Dn{32.6} & \Dc{34.4} \\
MoCo-v2-a & 66.4 & \Dn{28.1} & \Dn{40.5} & \Dn{29.4} & \Dn{36.8} & \Dn{29.4} & \Dc{32.8} \\
MoCo-v2-b & 71.1 & \Dn{31.3} & \Dn{45.2} & \Dn{31.0} & \Dn{39.7} & \Dn{31.3} & \Dc{35.7} \\
SimCLR-v2 & 71.0 & \Dn{31.5} & \Dn{45.4} & \Dn{30.8} & \Dn{43.0} & \Dn{31.7} & \Dc{36.5} \\
BarlowTwins & 73.5 & \Dn{26.7} & \Dn{39.8} & \Dn{29.7} & \Dn{43.0} & \Dn{34.4} & \Dc{34.7} \\
DeepCluster-v2 & 75.2 & \Dn{28.2} & \Dn{41.1} & \Dn{28.5} & \Dn{43.2} & \Dn{38.9} & \Dc{36.0} \\
SwAV-a & 72.0 & \Dn{27.0} & \Dn{39.8} & \Dn{28.3} & \Dn{40.6} & \Dn{33.9} & \Dc{33.9} \\
SwAV-b & 74.9 & \Dn{26.8} & \Dn{39.3} & \Dn{28.6} & \Dn{41.4} & \Dn{36.3} & \Dc{34.5} \\
\hline
\end{tabular}
}
\vspace{-10pt}
\end{table}

\section{Results}

From the downstream data corruption tests, we observe that CL is generally more robust than SL, although different CL algorithms show varied degrees of robustness. Some example corruptions are shown in Figure~\ref{fig:extreme}. However, when data corruption is applied to pre-training, the models' behaviors become complicated -- whether CL is more robust depends on the type of corruption.

\def\D#1{\scriptsize\textcolor{blue}{(#1\%)}}
\def\Dn#1{{#1\%}}

\def\MinNumber{20.0}
\def\MaxNumber{95.0}

\begin{table}[t]
\caption{Robustness Test I: \emph{downstream pixel- and patch-level} corruptions with \emph{ViT} backbone \cite{dosovitskiy2020image}. We show KNN accuracies and the $\Delta$'s on three datasets. Similar to Table~\ref{tab:downstream_resnet}, ViT CL models are also more robust than the two SL models, especially to gamma distortion. The generative method, MAE \cite{he2021masked}, is slightly more robust than CL to patch shuffling on CIFAR, but inferior on STL10 and more vulnerable to gamma distortion.}
\label{tab:downstream_vit}
\small
\centering
\setlength{\tabcolsep}{2pt}
\scalebox{0.6}{%
\begin{tabular}{@{}lcccccccc@{}}
\hline
STL10 & Orig & $\gamma 0.2$ & $\gamma 2.5$ & G4x4 & G24x24 & L4x4 & L24x24  & Avg $\Delta$ \\
\hline
ViT (Sup)  & 98.85 & 91.71 \D{7.2} & 91.39 \D{7.5} & 88.96 \D{10.0} & 43.69 \D{55.8} & 45.95 \D{53.5} & 70.89 \D{28.3}  & \Dc{27.1} \\
DeiT (Sup) & 98.64 & 97.58 \D{1.1} & 98.01 \D{0.6} & 92.92 \D{5.8} & 46.99 \D{52.4} & 45.60 \D{53.8} & 73.22 \D{25.8}  & \Dc{23.3} \\
DINO     & 98.91 & 98.31 \D{0.6} & 98.17 \D{0.7} & 95.30 \D{3.7} & 50.36 \D{49.1} & 52.35 \D{47.1} & 79.96 \D{19.2}  & \Dc{20.1} \\
MoCo-v3  & 97.89 & 97.11 \D{0.8} & 96.75 \D{1.2} & 91.24 \D{6.8} & 48.86 \D{50.1} & 47.70 \D{51.3} & 74.88 \D{23.5}  & \Dc{22.3}  \\
MAE      & 90.74 & 83.54 \D{7.9} & 87.42 \D{3.7} & 72.54 \D{20.1} & 46.35 \D{48.9} & 46.20 \D{49.1} & 60.15 \D{33.7}  & \Dc{27.2}  \\
\hline
CIFAR10 & Orig & $\gamma 0.2$ & $\gamma 2.5$ & G4x4 & G8x8 & L4x4 & L8x8  & Avg $\Delta$ \\
\hline
ViT (Sup) & 94.23 & 71.42 \D{24.2} & 82.37 \D{12.6} & 64.09 \D{32.0} & 52.58 \D{44.2} & 52.54 \D{44.2} & 59.63 \D{36.7}  & \Dc{32.3}  \\
DeiT (Sup) & 95.37 & 90.66 \D{4.9} & 92.78 \D{2.7} & 73.24 \D{23.2} & 59.48 \D{37.6} & 53.10 \D{44.3} & 59.65 \D{37.5}  & \Dc{25.0}  \\
DINO & 96.68 & 92.85 \D{4.0} & 94.65 \D{2.1} & 77.99 \D{19.3} & 64.63 \D{33.2} & 60.79 \D{37.1} & 68.04 \D{29.6}  & \Dc{20.9}  \\
MoCo-v3 & 96.16 & 91.90 \D{4.4} & 94.17 \D{2.1} & 75.30 \D{21.7} & 61.14 \D{36.4} & 57.60 \D{40.1} & 64.43 \D{33.0}  & \Dc{22.9}  \\
MAE & 77.06 & 71.00 \D{7.9} & 72.04 \D{6.5} & 61.25 \D{20.5} & 55.06 \D{28.5} & 53.31 \D{30.8} & 56.99 \D{26.0}  & \Dc{20.0}  \\
\hline
CIFAR100 & Orig & $\gamma 0.2$ & $\gamma 2.5$ & G4x4 & G8x8 & L4x4 & L8x8  & Avg $\Delta$ \\
\hline
ViT (Sup) & 79.86 & 48.70 \D{39.0} & 60.87 \D{23.8} & 40.95 \D{48.7} & 29.84 \D{62.6} & 28.91 \D{63.8} & 35.31 \D{55.8}  & \Dc{49.0}  \\
DeiT (Sup) & 78.23 & 68.98 \D{11.8} & 73.00 \D{6.7} & 49.86 \D{36.3} & 34.81 \D{55.5} & 29.49 \D{62.3} & 36.12 \D{53.8}  & \Dc{37.7}  \\
DINO & 83.88 & 75.76 \D{9.7} & 79.21 \D{5.6} & 56.81 \D{32.3} & 40.80 \D{51.4} & 36.62 \D{56.3} & 44.82 \D{46.6}  & \Dc{33.7} \\
MoCo-v3 & 82.32 & 73.25 \D{11.0} & 77.42 \D{6.0} & 53.07 \D{35.5} & 37.75 \D{54.1} & 33.00 \D{59.9} & 40.79 \D{50.4}  & \Dc{36.2}  \\
MAE & 53.70 & 47.72 \D{11.1} & 49.46 \D{7.9} & 37.18 \D{30.8} & 30.93 \D{42.4} & 29.36 \D{45.3} & 34.18 \D{36.4}  & \Dc{29.0}  \\
\hline
\end{tabular}
}
\vspace{-10pt}
\end{table}

\subsection{CL is more robust to downstream data corruption than SL}

We conduct downstream robustness tests on various datasets with frozen ResNet-50 \cite{he2016deep} in Table~\ref{tab:downstream_resnet} and ViT \cite{dosovitskiy2020image} in Table~\ref{tab:downstream_vit}. The model checkpoints are obtained from VISSL \cite{goyal2021vissl} and the official code bases. They are pre-trained on the clean version of ImageNet. We employ pixel-level and patch-level corruptions and report KNN accuracy. The raw accuracy numbers are in Appendix \ref{sec:b.3}.

For pixel-level corruption, we pick gamma distortion with $\gamma=\{0.2, 5\}$. For patch-level corruption, we choose one small patch size and one large patch size, for local and global shuffling each. In both tables, the CL methods have demonstrated higher robustness (lower average $\Delta$) than the SL method.
The same observation holds if we unfreeze the backbone and fine-tune fully with an additional linear layer as shown in Appendix \ref{sec:b.4}.

Interestingly, not all CL methods are equally robust; even within the same method, models trained with different hyper-parameters (such as epochs) exhibit different levels of robustness (\eg, comparing SwAV-a and SwAV-b). With ResNet-50, we notice SimSiam, SwAV, and BarlowTwins to behave slightly more robust than others.

\def\MinNumber{10.0}
\def\MaxNumber{60.0}

\begin{table}[t]
\caption{Robustness Test II: \emph{pre-training pixel- and patch-level} corruptions of CIFAR10, with ResNet18 backbone and KNN evaluation. We pre-train four CL methods and SL with $p=\{4,8\}$ patch and $\gamma=0.2$ corruptions, and discover that SL is more robust than CL in this scenario. While CL methods obtain average $\Delta$ above $20\%$, SL achieves $16.7\%$, which is lower than the best CL method here (MoCo v2 \cite{chen2020improved}).}
\label{tab:pretrain_c10}
\centering
\footnotesize
\setlength{\tabcolsep}{2pt}
\scalebox{0.73}{%
\begin{tabular}{@{}lccccccc@{}}
\hline
CIFAR10  & Orig  & $\gamma 0.2$ & G4x4 & G8x8 & L4x4 & L8x8  & Avg $\Delta\downarrow$  
\\
\hline
Sup & 89.53 & 87.36 \D{2.4} & 76.06 \D{15.0} & 65.88 \D{26.4} & 65.94 \D{26.3} & 77.49 \D{13.4} & \Dc{16.7} \\
MoCo-v2 & 88.73 & 85.84 \D{3.3} & 67.18 \D{24.3} & 60.51 \D{31.8} & 63.35 \D{28.6} & 76.90 \D{13.3} & \Dc{20.3} \\
BYOL & 88.39 & 82.72 \D{6.4} & 67.47 \D{23.7} & 60.63 \D{31.4} & 62.64 \D{29.1} & 75.15 \D{15.0} & \Dc{21.1} \\
Barlow & 88.89 & 80.49 \D{9.4} & 68.34 \D{23.1} & 61.13 \D{31.2} & 62.53 \D{29.7} & 75.28 \D{15.3} & \Dc{21.7} \\
DINO & 84.75 & 69.27 \D{18.3} & 64.26 \D{24.2} & 55.83 \D{34.1} & 58.57 \D{30.9} & 68.96 \D{18.6} & \Dc{25.2} \\
\hline
\end{tabular}
}
\vspace{-10pt}
\end{table}

\def\D#1{\\\textcolor{blue}{(#1\%)}}
\def\MinNumber{0.0}
\def\MaxNumber{60.0}

\begin{table}[t]
\caption{Robustness Test II: \emph{pre-training pixel- and patch-level} corruptions of ImageNet100. We focus our comparison on MoCo-v2 and SL to train on corrupted ImageNet100, which is a 100-class subset of ImageNet and substantially larger than CIFAR. SL still shows higher robustness to our pixel-level and patch-level corruptions, in agreement with Table~\ref{tab:pretrain_c10}.}
\label{tab:pretrain_in100}
\centering
\footnotesize
\setlength{\tabcolsep}{2pt}
\scalebox{0.7}{%
\begin{tabular}{@{}llccccccccc@{}}
	\hline
	IN-100 & Orig & $\gamma0.2$ & $\gamma5$ & G2x2 & G4x4 & G8x8 & L128x128 & L64x64 & L32x32 
	& Avg $\Delta\downarrow$
	\\
	\hline
	Sup & 77.08 
	& \makecell{73.60 \D{4.5}}
	& \makecell{70.28 \D{8.8}}
	& \makecell{67.26 \D{12.7}}
	& \makecell{62.84 \D{18.5}}
	& \makecell{58.20 \D{24.5}}
	& \makecell{75.28 \D{2.3}}
	& \makecell{72.68 \D{5.7}}
	& \makecell{68.52 \D{11.1}}
	& \Dc{6.65}
	\\
	MoCo-v2 & 74.38 
	& \makecell{66.80 \D{10.2}}
	& \makecell{62.74 \D{15.6}}
	& \makecell{44.90 \D{39.6}}
	& \makecell{35.84 \D{51.8}}
	& \makecell{30.94 \D{58.4}}
	& \makecell{69.34 \D{6.8}}
	& \makecell{63.68 \D{14.4}}
	& \makecell{54.24 \D{27.1}}
	& \Dc{28.0}
	\\
	\hline
\end{tabular}
}
\vspace{-10pt}
\end{table}

\subsection{CL and SL's robustness to pre-training data corruption depends on the corruption type}

During pre-training robustness tests, whether CL is more robust than SL depends on the type of corruption. We conduct pixel-level and patch-level corruption tests on CIFAR-10 shown in Table~\ref{tab:pretrain_c10} and on ImageNet-100 in Table~\ref{tab:pretrain_in100}. We then study dataset-level corruption on CIFAR-10 and ImageNet-LT. In our experiments, we find that SL is more robust to pixel- and patch-level corruptions, while CL is more robust to dataset-level corruptions.

\smallskip\paranoind{Pixel-Level and Patch-Level Corruption.}
Table~\ref{tab:pretrain_c10} shows the impacts of gamma distortion and patch global/local shuffling on CL and SL during pre-training. For each algorithm, we train for the same number of epochs with different corruption parameters and calculate $\Delta(\text{Alg})$ correspondingly. We train SL for 30 epochs and CL for 200 epochs (except for DINO which is trained for 600 epochs), since their clean data accuracy roughly matches each other. The accuracy is obtained by linear evaluation.
The drop of accuracy of SL due to gamma distortion is $2.4\%$, which is smaller than all the tested CL methods.
As for pre-training patch shuffling corruption, all CL methods behave similarly and less robustly than SL, except for the L8x8 case where Sup and MoCo-v2 are comparable. Appendix \ref{sec:b.2} reports the same observation with more training epochs. Appendix \ref{sec:b.1} shows that MoCo-v3 \cite{chen2021empirical} with ViT backbone is also less robust than SL to patch shuffle on CIFAR10.

The same observation carries over to the larger-scale ImageNet-100 experiment result in Table~\ref{tab:pretrain_in100}, where we compare MoCo-v2 (since it appears to be the most robust on CIFAR-10) and Sup in a total of 8 corruption settings. MoCo-v2 on average yields a $28.0\%$ degradation, while Sup only degrades $6.65\%$.

\def\D#1{\textcolor{blue}{(#1\%)}}

\begin{table}[t]
\caption{Robustness Test II: \emph{pre-training synthesized data substitution}. C10 and C100 refer to CIFAR-10 and CIFAR-100, respectively. Interestingly, at absolute scale, MoCo shows higher downstream transfer accuracy to CIFAR-100 than SL, even through the 10 pre-training classes are only a very small subset of the CIFAR-100 classes. In all three evaluation settings, MoCo-v2 demonstrates much more robustness (on average, $\Delta_{\text{MoCo}}=0.93\%$) than Sup ($\Delta_{\text{Sup}}=7.71\%$) to the distribution shift of synthesized data.}
\label{tab:pretrain_gan}
\centering
% \footnotesize
\small
\setlength{\tabcolsep}{2pt}
\scalebox{0.75}{
\begin{tabular}{lllllc}
\hline
Pre-train Alg & Pre-train Data & C10 Test  & C10 GAN Test  & C100 Test  & Avg $\Delta\downarrow$ \\
\hline
Sup & Orig C10 & 87.8  & 88.3  &  16.08  & - \\
    & GAN C10  & 80.0 \D{8.88}  & 82.8 \D{6.23}  &  14.79 \D{8.02} & \Dc{7.71} \\
MoCo-v2 & Orig C10 & 82.6  & 85.1  &  45.47  & - \\
    & GAN C10  & 82.2 \D{0.48}  & 85.4 \D{-0.35} &  44.27 \D{2.64}  & \Dc{0.93} \\
\hline
\end{tabular}
}
\vspace{-10pt}
\end{table}

\begin{table}[tb]
\caption{Robustness Test II: \emph{pre-training class imbalance}. We compare MoCo-v2 and SL (with ResNet-50) on a long-tail subset of ImageNet, ImageNet-LT \cite{liu2019large}, and a uniform subset we construct, ImageNet-UF. We use linear evaluation protocal: train a linear classifier on ImageNet-UF, and then report top-1 and different shot accuracy on ImageNet-LT-Val (20K images). Low-shot refers to classes with less than 20 images, many-shot the ones with more than 100, and med-shot in between. MoCo shows less sensitivity to pre-train data imbalance than Sup, as the drop in top-1 accuracy is smaller and the difference among shots is also small.}
\label{tab:pretrain_longtail}
\centering
\small
\setlength{\tabcolsep}{2pt}
\scalebox{0.74}{%
\begin{tabular}{@{}llllll@{}}
    \hline
    Alg & Pre-train Data & Top-1 & Low & Med & Many  \\
    \hline
    Sup & ImageNet-UF & 46.37 & 44.85 & 45.88 & 47.52 \\
        & ImageNet-LT & 44.90 \D{3.17} & 40.99 \D{8.61} & 43.48 \D{5.23} & 48.05 \D{-1.12} \\
    MoCo-v2 & ImageNet-UF & 32.36 & 30.63 & 31.66 & 33.84 \\
            & ImageNet-LT & 32.13 \D{0.71} & 30.99 \D{-1.18} & 31.45 \D{0.66} & 33.36 \D{1.42} \\
    \hline
\end{tabular}
}
\vspace{-10pt}
\end{table}

\smallskip\paranoind{Dataset-Level Corruption.}
We first investigate pre-train distribution shift caused by synthesized data. We adopt a class-conditional StyleGAN2-ADA\cite{karras2020training} generator trained on CIFAR-10 to synthesize another copy of CIFAR-10. The number of examples is kept the same as the original dataset. We train MoCo-v2 and the supervised model with different train/test data settings for different length of epochs. Table~\ref{tab:pretrain_gan} shows the performance differences due to synthesized data. When training on the synthesized data and testing on the original CIFAR-10, MoCo-v2 only has $2.58\%$ accuracy drop, greatly outperforming the supervised method which drops $8.44\%$. Evaluating on a GAN-synthesized test set yields similar observation -- MoCo-v2 shows almost no drop while Sup drops $6\%$. We also conduct downstream experiments for this corruption: testing the classification accuracy of the pre-trained representation on CIFAR-100. The results are also reported in Table~\ref{tab:pretrain_gan}. We observe that the drop of MoCo is also smaller than that of Sup. In our experiments, MoCo has shown more robustness on average to dataset-level corruption caused by an imperfect GAN.

Table~\ref{tab:pretrain_longtail} shows the impact of the other type of dataset-level corruption, the class imbalance. We use the ImageNet-LT (long-tail) dataset to simulate the real-world long-tail class distribution \cite{liu2019large}. For comparison, we sample a balanced subset of ImageNet, called ImageNet-UF (uniform), with the same total number of images as ImageNet-LT. Then we compare the recognition accuracy on the ImageNet-LT validation split of the fine-tuned linear classifiers on ImageNet-UF. The backbone is ResNet-50 \cite{he2016deep}. Although there is a gap between the baseline top-1 accuracy of MoCo-v2 and Sup, we observe that the decline of MoCo resulting from pre-training on the long-tail rather than the uniform version is much smaller than SL. In fact, the MoCo performance appears to be insensitive to class balance or imbalance (the top-1 $\Delta$ is only $0.71\%$). This is in contrary to SL, which shows a larger drop. The difference is more salient by looking at the low-shot ($<20$ images per class), medium-shot, and many-shot ($>100$ images per class) accuracy separately. Supervised pre-training on the long-tail version sacrifices the low-shot accuracy for a higher many-shot accuracy, whereas MoCo-v2 pre-training shows insignificant difference among the shots. Our observation is consistent with a contemporary work \cite{liu2021self}.

\subsection{Discussion}

We try to balance diversity and setup unity under compute budget. Within each table, the setup is \emph{consistent}, allowing comparison of SL and CL; across tables, we intentionally evaluate if the observation is \emph{generalizable} across backbones and datasets. For example, Tables~\ref{tab:downstream_resnet} and~\ref{tab:downstream_vit} are the \emph{same} corruptions but varying backbones; Table~\ref{tab:pretrain_in100} extends the \emph{same} observation from small-scale in Table~\ref{tab:pretrain_c10} to larger-scale data.

The $\Delta$ metric can become unreliable when the original uncorrupted accuracy differs too much across methods. We overcame it by: (1) controlling the original accuracy to be relatively close, \eg, in Table~\ref{tab:downstream_resnet} the `b' version has closer ImageNet accuracy to SL, and Tables~\ref{tab:pretrain_c10} and~\ref{tab:pretrain_in100} have comparable original accuracy; (2) testing multiple datasets, backbones, and corruption settings to draw consistent conclusions from more data points.

\section{Analysis}

\subsection{Strong dependency on data augmentation may explain CL's non-robustness to patch shuffling}

The different degree of dependency on data augmentation of CL and SL may explain why CL algorithms are less robust to pixel-level and patch-level pre-training corruptions. Contrastive learning relies heavily on well-defined data augmentation, while supervised learning can train without data augmentation as Row 1 of Table~\ref{tab:aug}. The central assumption in using data augmentation is that the augmented image falls within or close to the natural image statistics, \eg, a random cropped image is still plausible. However, our data corruption such as patch shuffling destroys the structure of an image and thus renders the random resize-crop augmentation inappropriate, as the cropped image no longer has the same global structure.

To support our claim, we switch the order of the patch shuffling corruption and the standard augmentation in the MoCo experiments in Table~\ref{tab:aug} (KNN evaluation), since shuffling after crop gives consistent image structure. We find that this switching reverts much of the accuracy drop, making MoCo comparably robust to Sup.

\begin{table}[tb]
\small
\centering
\caption{Additional pre-training corruption with Sup no-augmentation and Sup/MoCo augmentation-then-corrupt variants. SL is able to learn without data augmentation. Contrary to the corrupt-aug version in previous sections, MoCo and Sup share roughly a similar level of robustness with the aug-corrupt version.}
\setlength{\tabcolsep}{2pt}
\scalebox{0.65}{%
\begin{tabular}{@{}lcccccc@{}}
\hline
Pre-training & Orig. & G4x4 & G8x8 & L4x4 & L8x8 & $\gamma0.1$\\
\hline
Sup no-aug       & 87.66 & 77.37 \D{11.7} & 71.86 \D{18.0} & 73.30 \D{16.4} & 82.34 \D{6.0}  & 86.86 \D{0.91} \\
Sup aug-corrupt  & 92.23 & 85.92 \D{ 6.8}  & 80.58 \D{12.6} & 83.61 \D{ 9.4}  & 89.96 \D{2.5}  & - \\
MoCo corrupt-aug & 82.55 & 65.43 \D{17.1} & 59.49 \D{27.9} & 59.62 \D{27.8} & 70.14 \D{15.}  & - \\
MoCo aug-corrupt & 82.55 & 77.63 \D{ 6.0}  & 73.48 \D{11.0} & 78.12 \D{ 5.4}  & 81.25 \D{1.6}  & - \\
\hline
\end{tabular}
}
\label{tab:aug}
\vspace{-6pt}
\end{table}

\subsection{CL and SL behave differently during training}

\begin{figure}[t]
  \centering
  \includegraphics[width=0.98\linewidth]{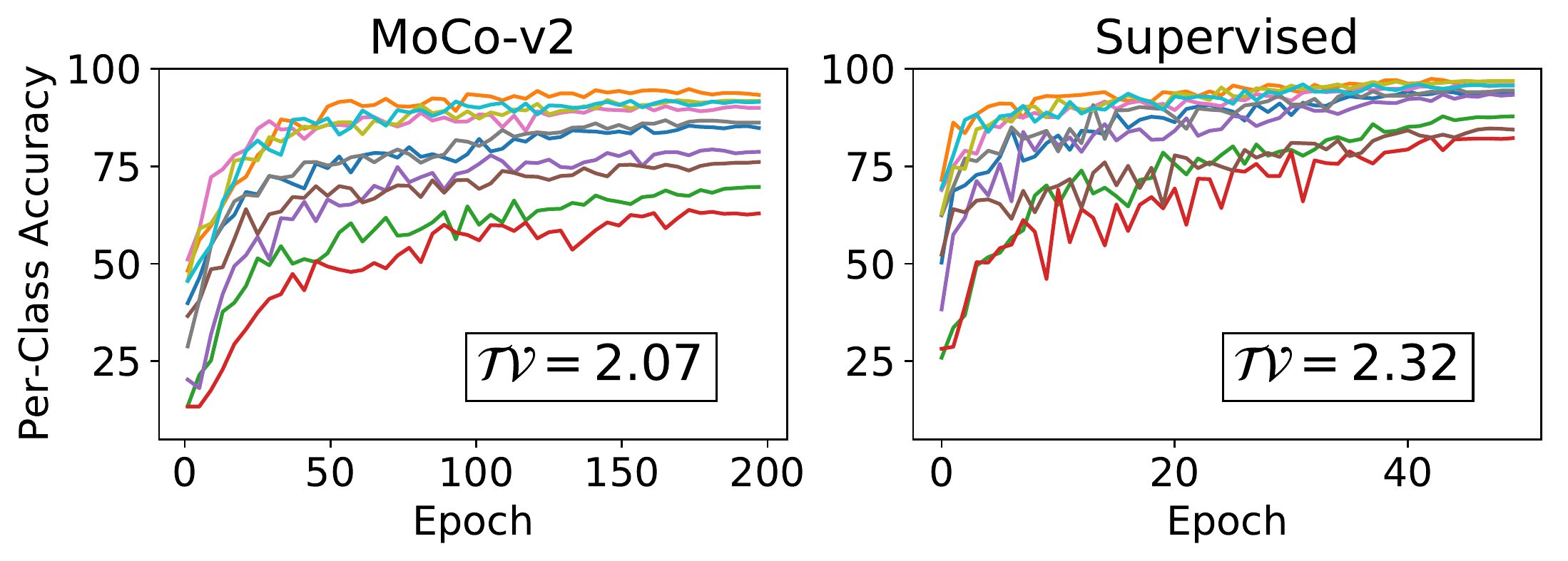}
  \caption{Class-wise test accuracy (\ie, recall) of MoCo and SL on original CIFAR-10 during training. MoCo has more steady class-wise accuracy curves and smaller mean feature semantic fluctuation ($\mathcal{TV}$) than SL.}
  \label{fig:acc}
  \vspace{-7pt}
\end{figure}

The robustness discrepancy between CL (\eg, MoCo) and SL is not only reflected in the final trained models, but is in fact also attributed in the training process. To analyze how the feature space evolves during training, we measure the following three metrics.

\smallskip\paranoind{Feature Semantic Fluctuation.} We can monitor the classification ability of the feature extractor by the accuracy of a KNN probe. We define feature semantic fluctuation of class $i$ as the total variation of per-class accuracy of class $i$ (as a function of epoch $t$) averaged over all epochs:
$
    \mathcal{TV}_i = \frac1{T-1} \sum_{t=0}^{T-2} |\text{Acc}^{(i)}_{t+1} - \text{Acc}^{(i)}_t|.
    \label{eq:tv}
$
We further define the mean feature semantic fluctuation as the mean of $\mathcal{TV}_i$ over all classes. Larger semantic fluctuation indicates less stable feature space.

\smallskip\paranoind{Feature Uniformity.} We can measure the uniformity of all the features or class-wise features as the log-mean of Gaussian potentials of the normalized features:
$
U(f_t,\mathcal{D})=
-\log \mathbb{E}_{x_0, x_1\sim \mathcal{D}} \left[e^{-2\|f_t(x_0) - f_t(x_1)\|_{2}^{2}}\right].
$
Here $f_t$ is the network at epoch $t$, $\mathcal{D}$ is the dataset, and $x_0$ and $x_1$ are images sampled from the dataset. The use of this measure to study contrastive learning is exemplified in \cite{wang2020understanding}. Intuitively, a greater $U$ means more uniformly distributed features on the unit sphere, while a smaller value means more concentrated features.

\smallskip\paranoind{Feature Distance.} We also measure the average feature squared $\ell_2$ distance between two classes. A larger distance could mean more linear separability. Denoting $\mathcal{D}_i$ and $\mathcal{D}_j$ as feature matrices of two classes, the feature distance is calculated as:
$
d(f_t, \mathcal{D}_i, \mathcal{D}_j) = \mathbb{E}_{x_0\sim\mathcal{D}_i, x_1\sim \mathcal{D}_j}
\left[\|f_t(x_0) - f_t(x_1)\|_{2}^2\right].
$
Note that if $\mathcal{D}_i = \mathcal{D}_j$, it actually measures the intra-class variance of class $i$.

\begin{figure}[t]
  \centering
  \includegraphics[width=0.98\linewidth]{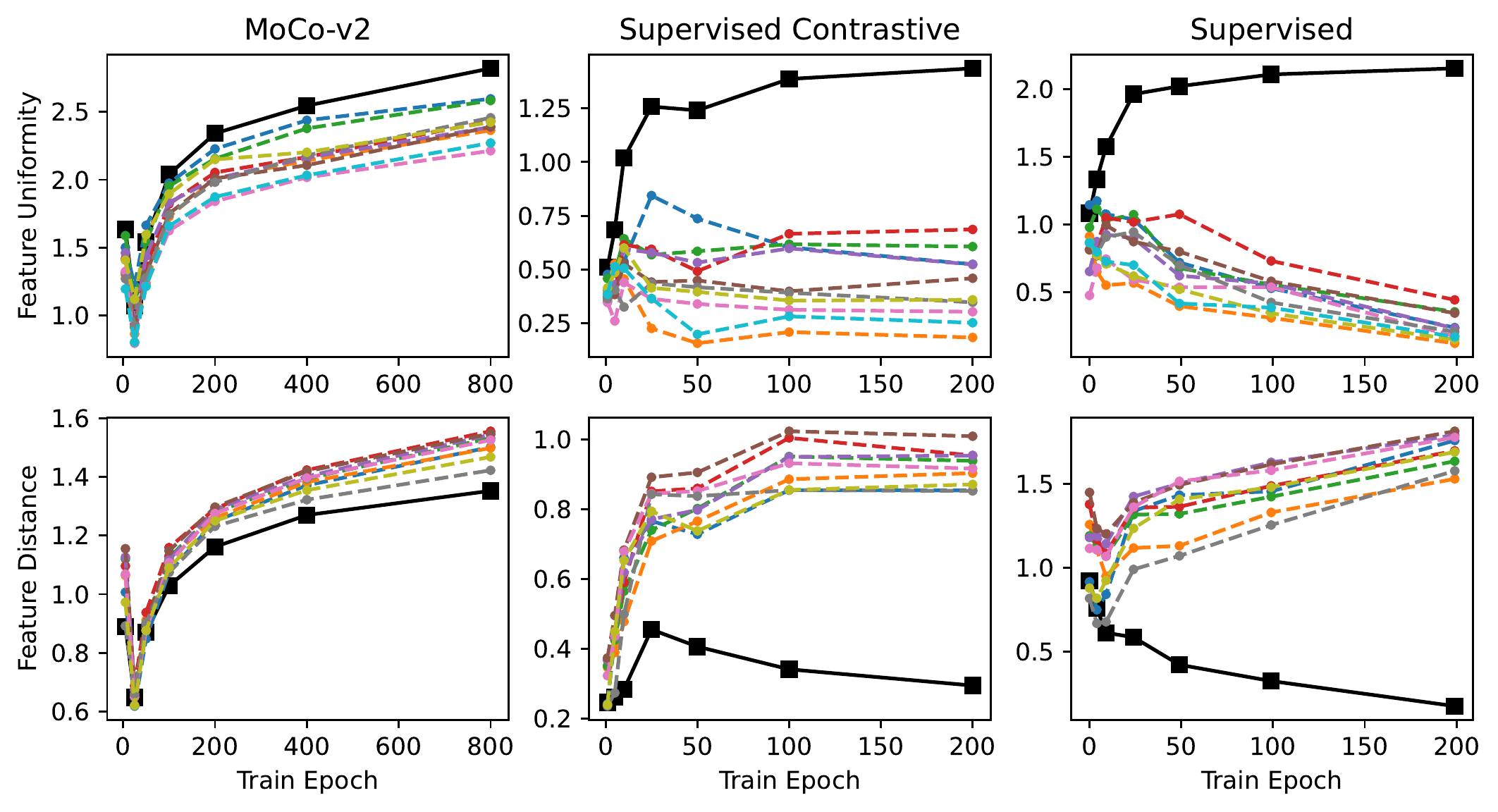}
  \caption{\textbf{Above:} Solid black line refers to the uniformity of the overall feature space. Dashed lines refer to the class-wise feature uniformities of the 10 classes. While the overall uniformity of all methods grows, the uniformity of each class (related to intra-class variance) of Sup or SupCon is shrinking as training progresses. In the end, the overall uniformity of MoCo is the largest.
  \textbf{Below:} Solid black line refers to $d(f_t, \mathcal{D}_0, \mathcal{D}_0)$, \ie, the intra-class variance of class 0. Dashed lines refer to feature distances between $\mathcal{D}_i (i\ne 0)$ and $\mathcal{D}_0$. While the distances between classes increase in all methods, the intra-class variance behavior of MoCo (increasing) is the opposite to that of Sup or SupCon (decreasing).}
 \label{fig:unifdist}
 \vspace{-10pt}
\end{figure}

We train ResNet-18\cite{he2016deep} on the original CIFAR-10\cite{krizhevsky2009learning} train split and measure the above metrics on the test split. Figure \ref{fig:unifdist} shows the dynamics of feature uniformity and distances of MoCo-v2 \cite{he2020momentum,chen2020improved}, supervised contrastive (SupCon) \cite{khosla2020supervised}, and supervised learning. We are interested in SupCon, because it bridges CL and SL by leveraging a similar contrastive loss.

As illustrated, the overall feature uniformity of MoCo-v2\cite{chen2020improved} is greater than 2.5 and approaching 3, while the overall feature uniformity of SupCon and supervised methods range from 1.25 to 2.2. This means that features from contrastive learning methods are more uniformly distributed on the unit sphere. 
By looking at the class-wise feature uniformity and distance, we notice that the supervised model tends to compress (and maybe over-compress) the features of each class. Figure~\ref{fig:acc} shows that the accuracy of a KNN probe during supervised learning also fluctuates more dramatically.
We can interpret it as that the classes are competing with each other, and SL cannot improve the performance on all classes at the same time like CL methods.

\subsection{Uniformity regularization improves downstream robustness of supervised pre-training}

\begin{table}[t]
\centering
\setlength{\tabcolsep}{4.9pt}
\caption{Supervised pre-training with uniformity regularization improves test-time robustness (ResNet-18, CIFAR-10, 200 epochs). The model achieves higher KNN evaluation accuracy on corrupted data while sacrificing little accuracy on the original data. Subtracting the uniformity promoting term appears to have the opposite effect.}
\footnotesize
\scalebox{0.85}{%
\begin{tabular}{@{}llccccc@{}}
\hline
Pre-train Alg   & Metric  & Orig & $\gamma5$ & L4x4  & G4x4  \\
\hline
Sup            & Acc, Unif  & 94.18, 1.98  & 72.85, 1.64 & 37.85, 0.98  & 39.70, 0.90    \\
Sup$+0.01$Unif & Acc, Unif  & 94.21, 2.69  & 74.47, 2.03 & 42.22, 1.11  & 44.34, 1.30    \\
Sup$-0.01$Unif & Acc, Unif  & 94.56, 1.12  & 71.50, 0.77 & 36.15, 0.41  & 37.88, 0.46    \\
\hline
\end{tabular}
}
\label{tab:supunif}
\vspace{-10pt}
\end{table}

The analysis above shows that MoCo appears to yield a more uniformly distributed feature space and tends not to compress the intra-class variance of semantic classes. Could the larger uniformity be one reason behind the higher downstream robustness of MoCo? We introduce a small uniformity-promoting regularization term in addition to the cross-entropy loss in SL. This regularization is computed from the mini-batch itself without using the Siamese architecture or memory bank in MoCo.

In Table~\ref{tab:supunif}, there is no surprise that adding (or subtracting) the uniformity regularization produces a more (or less) uniform test feature space. We also notice a correlation between the KNN evaluation accuracy and the test set uniformity under 3 corruption conditions. This experiment suggests that we could improve SL by leveraging loss functions from CL and potentially get the best of both worlds.

\section{Conclusion}

This paper systematically studies the robustness of contrastive learning and supervised learning through a diverse set of multi-level data corruption robustness tests. We discover interesting robustness behaviors of contrastive learning to different corruption settings. We attempt to explain the observations through augmentation and feature space analyses. Our analysis of the feature learning dynamics gives insight that uniformity might be the key to higher downstream robustness. Our results favor the current use of CL or a combination of CL and SL in visual representation pre-training.

\smallskip\paranoind{Limitation and Future Work.}
The scale of some experiments is not large.
While our exploration with small/mid-scale data already reveals novel and consistent behavior differences, larger-scale (such as full ImageNet) experiments and even more diverse settings would be ideal and left for future work.
We observe some opposite robustness behaviors between pre-training and downstream, and hypothesize two initial explanations for them (interference with augmentation and feature uniformity, respectively). As shuffling can not be bounded well by the $\ell_p$ input perturbation formulation in adversarial robustness, establishing theory is hard. A recent work formalizing coordinate transforms \cite{balestriero2022data} might be useful for developing theory. Our derived robustness tests and metrics may have standalone value in other contexts such as model selection. We will continue working in this direction and hope our study can motivate the community to further explore these behavioral differences of learning machinery.

%%%%%%%%% REFERENCES
{\small
\bibliographystyle{ieee_fullname}
\bibliography{egbib}
}

\appendix
\renewcommand{\thefigure}{\Alph{section}.\arabic{subsection}}

\renewcommand{\thetable}{\Alph{section}.1}

\section{Additional Implementation Details}

Table~\ref{tab:config} below lists the experiment configurations for each pre-training robustness table of the main paper. 
We train our own ResNets \cite{he2016deep} on CIFAR-10 \cite{krizhevsky2009learning} and ImageNet variants \cite{deng2009imagenet}. ImageNet-LT/UF are the long-tail and uniformly-subsampled versions. ImageNet-100 is a 100 class subset of full ImageNet-1K. We mainly list Sup and MoCo-v2 \cite{chen2020improved} hyper-parameters here. The other CL methods follow their recommended hyper-parameter values in the Solo-Learn package \cite{JMLR:v23:21-1155}.

\begin{table}[ht]
\caption{Implementation details for the pre-training results in the main paper.}
    \centering
    \footnotesize
    \setlength{\tabcolsep}{4pt}
    \begin{tabular}{l|ccc}
    \hline
    Config       & Tables~\ref{tab:pretrain_c10},\ref{tab:pretrain_gan}  & Table~\ref{tab:pretrain_longtail}  & Table~\ref{tab:pretrain_in100}  \\
    \hline
    Pre-train dataset& CIFAR-10  & ImageNet-LT/UF & ImageNet-100  \\
    \# of categories      & 10        & 1000           & 100 \\
    Train image size& 32        & 224            & 224 \\
    Train data size & 50K      & 115K           & 130K \\ 
    Network         & ResNet-18 & ResNet-50      & ResNet-18    \\
    Backbone out dim& 512       & 2048           & 512 \\
    Sup epochs      & 50        & 200            & 50 \\
    Sup lr          & 0.1 cos   & 0.015 cos      & 0.015 cos \\
    Sup batch size  & 512       & 128            & 128 \\
    MoCo epochs     & 200       & 200            & 200 \\
    MoCo lr         & 0.06 cos  & 0.015 cos      & 0.03 cos \\
    MoCo batch size & 512       & 128            & 256 \\
    MoCo dim        & 128       & 128            & 128 \\
    MoCo temp.      & 0.1       & 0.2            & 0.2 \\
    MoCo momentum   & 0.99      & 0.999          & 0.999 \\
    MoCo queue size & 4096      & 65536          & 65536 \\
    Evaluation      & Linear & Linear  & Linear  \\
    Augmentation    & %\multicolumn{3}{c}{ResizeCrop+flip+color(.4,p=.8)+gray(p=.2)}
    \makecell{crop+flip+\\color(.4,p=.8)\\+gray(p=.2)} &
    \multicolumn{2}{c}{\makecell{crop+flip+\\color(.4,p=.8)+gray(p=.2)+\\gauss(.1,.2,p=.5)}}
    \\
    \hline
    \end{tabular}
    \label{tab:config}
\end{table}

\renewcommand{\thetable}{\Alph{section}.\arabic{subsection}}

\section{Additional Results}

\subsection{Pre-training robustness test with Transformer backbone}
\label{sec:b.1}

In the main paper, we compare pre-training robustness with a CNN backbone in Table~\ref{tab:pretrain_c10}, and show Vision Transformer (ViT) downstream robustness test results in Table~\ref{tab:downstream_vit}. Here, we supplement ViT \emph{pre-training} robustness test results. Specifically, we leverage MoCo-v3 \cite{chen2021empirical}, the ViT version of MoCo, and Supervised ViT. The results are in Table~\ref{tab:pretrain_vit}. We find that the MoCo-v3 degradation is larger with patch shuffling, but smaller with gamma distortion. Interestingly, the impact of patch shuffling is much smaller than a CNN (despite the Orig performance gap between ViT and CNN). We suspect that this is due to the unique patching and attention network structure of ViT. Essentially, if we do not take into consideration the data augmentation, with the right patch size, the shuffling within a small patch does not affect the learning of ViT much, and the global ordering of patches also does not matter much, because of learned positional embeddings and global attention.

\begin{table}[htb]
\caption{Pre-training robustness with ViT on CIFAR10: MoCo-v3 vs. Sup. For the ViT architecture, since the input size (32x32) is smaller than that of a standard ViT, we use a customized small ViT (image size=32, patch size=4, dim=512, depth=6, heads=8, mlp dim=512, dropout=0.1, emb dropout=0.1).}
\centering
\footnotesize
\setlength{\tabcolsep}{2.5pt}
\scalebox{0.9}{%
\begin{tabular}{l | c | c c c c c | c}
 \hline
 \textbf{Method}&Orig& G4x4 & G16x16 & L4x4 & L16x16 & Avg $\Delta$ & $\gamma=0.1$\\
 \hline
 Sup ViT 50ep  &67.92& 59.01 & 47.97 & 57.76 & 67.95 & - & 52.96\\
 $\Delta$   & - & 13.12\% & 29.37\% & 14.96\% & -0.04\% & 14.35\% & 22.03\% \\
 \hline
 MoCo-v3 200ep  &62.78& 53.36 & 41.58 & 53.52 & 61.77 & - & 51.41\\
 $\Delta$   & - & 15.0\% & 33.77\% & 14.75\% & 1.61\% & 16.28\% & 18.11\% \\
 \hline
\end{tabular}
}
\label{tab:pretrain_vit}
\end{table}

\subsection{Pre-training robustness test with longer epochs}
\label{sec:b.2}

In Table~\ref{tab:pretrain_c10} of the main paper, we mostly report results of short pre-training schedules: Sup 30 epochs and CL 200 epochs, in order to make the baseline results comparable. We report CIFAR-10 longer training epochs in Table~\ref{tab:pretrain_long}. Training longer does not change our observation that MoCo appears less robust to patch- and pixel-level corruptions than SL during pre-training on this dataset.

\begin{table}[htb]
\caption{Pre-training robustness: Sup 50ep vs. MoCo-v2 400ep, ResNet-18, CIFAR-10.}
\centering
\footnotesize
\setlength{\tabcolsep}{2.5pt}
\scalebox{0.9}{%
\begin{tabular}{l | c | c c c c c | c}
\hline
\textbf{Method}&Orig& G4x4 & G8x8 & L4x4 & L8x8 & Avg $\Delta$ & $\gamma=0.1$\\
\hline
Sup 50ep  &92.23& 81.14 & 71.33 & 71.72 & 81.95 & - & 89.11\\
$\Delta$   & - & 12.02\% & 22.67\% & 22.24\% & 11.15\% & 17.02\% & 3.38\% \\
\hline
MoCo-v2 400ep &91.43& 70.99 & 64.25 & 66.56 & 81.51 & - & 83.94\\
$\Delta$   & - & 22.36\% & 29.73\% & 27.20\% & 10.85\% & 22.54\% & 8.19\% \\
\hline
\end{tabular}
}
\label{tab:pretrain_long}
\end{table}

\subsection{Downstream robustness test with KNN: accuracy numbers}
\label{sec:b.3}

Table~\ref{tab:downstream_detail} shows the detailed accuracy numbers for computing the summary statistics in Table~\ref{tab:downstream_resnet} of the main paper.

\def\D#1{\scriptsize\textcolor{blue}{(#1\%)}}
\def\MinNumber{25.0}
\def\MaxNumber{95.0}
\def\Dc#1{%
	\pgfmathsetmacro{\PercentColor}{max(min(100.0*(#1 - \MinNumber)/(\MaxNumber-\MinNumber),100.0),0.00)} %
	\colorbox{yellow!\PercentColor}{#1\%}
}

\begin{table*}[p!]
\caption{Robustness to downstream data corruption with \emph{KNN evaluation}. This table contains the detailed top-1 accuracy numbers constituting Table~\ref{tab:downstream_resnet} in the main paper. The shades of yellow in the last column indicate the size of the numbers.}
\footnotesize
\centering
\setlength{\tabcolsep}{3pt}
\scalebox{0.95}{%
\begin{tabular}{lllllllll c}
\hline
Pre-train Alg & Dataset & Orig & $\gamma=0.2$ &$\gamma=5$ & G-small & G-large & L-small & L-large & Avg $\Delta$ \\
\hline
Sup	&	cifar10   	&	86.4	&	76.4 \D{11.7}	&	67.6 \D{21.8}	&	61.2 \D{29.2}	&	49.6 \D{42.6}	&	46.9 \D{45.7}	&	53.6 \D{38.0}	&	\Dc{31.5} \\
Sup	&	cifar100  	&	65.1	&	52.4 \D{19.6}	&	44.1 \D{32.3}	&	37.4 \D{42.6}	&	25.7 \D{60.5}	&	23.4 \D{64.1}	&	30.8 \D{52.8}	&	\Dc{45.3} \\
Sup	&	stl10     	&	96.6	&	92.2 \D{4.6}	&	82.6 \D{14.5}	&	80.8 \D{16.3}	&	40.7 \D{57.8}	&	43.4 \D{55.1}	&	60.3 \D{37.5}	&	\Dc{31.0} \\
Sup	&	cars196   	&	26.8	&	23.3 \D{13.2}	&	18.1 \D{32.5}	&	12.7 \D{52.7}	&	4.4 \D{83.6}	&	4.1 \D{84.9}	&	16.0 \D{40.4}	&	\Dc{51.2} \\
Sup	&	aircraft70	&	40.3	&	37.9 \D{6.0}	&	38.8 \D{3.6}	&	25.2 \D{37.4}	&	8.0 \D{80.3}	&	10.0 \D{75.2}	&	25.5 \D{36.7}	&	\Dc{39.9} \\
% imagenet ref: 76.1  &  avg: \Dc{39.8}
\hline
Sup-b	&	cifar10   	&	84.9	&	77.0 \D{9.3}	&	68.9 \D{18.9}	&	57.7 \D{32.0}	&	46.5 \D{45.2}	&	44.0 \D{48.2}	&	51.9 \D{38.9}	&	\Dc{32.1} \\
Sup-b	&	cifar100  	&	63.2	&	53.3 \D{15.7}	&	45.4 \D{28.1}	&	33.0 \D{47.8}	&	21.4 \D{66.1}	&	18.9 \D{70.0}	&	28.0 \D{55.7}	&	\Dc{47.2} \\
Sup-b	&	stl10     	&	96.0	&	92.8 \D{3.3}	&	83.6 \D{12.9}	&	77.9 \D{18.9}	&	36.0 \D{62.5}	&	41.5 \D{56.7}	&	60.4 \D{37.0}	&	\Dc{31.9} \\
Sup-b	&	cars196   	&	28.8	&	26.8 \D{7.0}	&	22.0 \D{23.8}	&	12.1 \D{58.1}	&	1.8 \D{93.7}	&	2.4 \D{91.6}	&	15.8 \D{45.2}	&	\Dc{53.2} \\
Sup-b	&	aircraft70	&	46.9	&	47.0 \D{-0.3}	&	46.0 \D{1.9}	&	29.8 \D{36.4}	&	7.7 \D{83.5}	&	10.8 \D{76.9}	&	29.8 \D{36.5}	&	\Dc{39.2} \\
% imagenet ref: 75.5  &  avg: \Dc{40.7}
\hline
BYOL	&	cifar10   	&	87.5	&	80.3 \D{8.3}	&	72.4 \D{17.3}	&	64.0 \D{26.8}	&	50.7 \D{42.1}	&	48.4 \D{44.7}	&	55.6 \D{36.5}	&	\Dc{29.3} \\
BYOL	&	cifar100  	&	67.4	&	58.1 \D{13.8}	&	49.8 \D{26.0}	&	39.7 \D{41.1}	&	26.1 \D{61.3}	&	24.5 \D{63.6}	&	32.1 \D{52.3}	&	\Dc{43.0} \\
BYOL	&	stl10     	&	94.9	&	92.4 \D{2.6}	&	85.0 \D{10.4}	&	78.1 \D{17.7}	&	41.5 \D{56.3}	&	43.8 \D{53.8}	&	63.0 \D{33.6}	&	\Dc{29.0} \\
BYOL	&	cars196   	&	22.5	&	22.3 \D{0.8}	&	18.7 \D{17.0}	&	12.6 \D{44.1}	&	2.5 \D{88.9}	&	2.3 \D{89.6}	&	18.6 \D{17.2}	&	\Dc{42.9} \\
BYOL	&	aircraft70	&	38.2	&	39.5 \D{-3.3}	&	38.6 \D{-1.2}	&	24.0 \D{37.2}	&	8.0 \D{79.2}	&	10.5 \D{72.6}	&	31.1 \D{18.5}	&	\Dc{33.8} \\
% imagenet ref: 72.3  &  avg: \Dc{35.6}
\hline
SimSiam	&	cifar10   	&	83.6	&	77.6 \D{7.2}	&	68.7 \D{17.8}	&	61.5 \D{26.5}	&	50.4 \D{39.8}	&	48.4 \D{42.2}	&	55.6 \D{33.5}	&	\Dc{27.8} \\
SimSiam	&	cifar100  	&	59.9	&	52.7 \D{12.1}	&	44.8 \D{25.1}	&	36.1 \D{39.7}	&	24.8 \D{58.6}	&	23.2 \D{61.2}	&	31.1 \D{48.0}	&	\Dc{40.8} \\
SimSiam	&	stl10     	&	92.1	&	89.6 \D{2.7}	&	81.1 \D{12.0}	&	74.2 \D{19.4}	&	39.1 \D{57.6}	&	43.8 \D{52.5}	&	62.7 \D{31.9}	&	\Dc{29.3} \\
SimSiam	&	cars196   	&	17.1	&	15.9 \D{6.9}	&	13.9 \D{18.5}	&	10.3 \D{39.4}	&	2.0 \D{88.6}	&	2.5 \D{85.2}	&	15.3 \D{10.2}	&	\Dc{41.5} \\
SimSiam	&	aircraft70	&	32.8	&	34.1 \D{-3.7}	&	32.6 \D{0.8}	&	20.9 \D{36.3}	&	6.8 \D{79.3}	&	10.8 \D{67.2}	&	27.6 \D{15.9}	&	\Dc{32.6} \\
% imagenet ref: 68.3  &  avg: \Dc{34.4}
\hline
MoCo	&	cifar10   	&	81.4	&	74.0 \D{9.1}	&	66.1 \D{18.9}	&	60.2 \D{26.0}	&	49.5 \D{39.2}	&	47.2 \D{42.0}	&	54.0 \D{33.7}	&	\Dc{28.1} \\
MoCo	&	cifar100  	&	56.6	&	48.2 \D{14.8}	&	41.9 \D{26.0}	&	35.3 \D{37.7}	&	23.5 \D{58.5}	&	23.1 \D{59.3}	&	30.0 \D{46.9}	&	\Dc{40.5} \\
MoCo	&	stl10     	&	90.1	&	88.1 \D{2.2}	&	77.5 \D{13.9}	&	73.5 \D{18.5}	&	39.9 \D{55.7}	&	42.8 \D{52.5}	&	60.0 \D{33.4}	&	\Dc{29.4} \\
MoCo	&	cars196   	&	13.1	&	12.8 \D{2.5}	&	11.3 \D{14.2}	&	8.8 \D{32.9}	&	2.3 \D{82.6}	&	2.6 \D{80.5}	&	12.1 \D{8.2}	&	\Dc{36.8} \\
MoCo	&	aircraft70	&	25.2	&	26.7 \D{-6.1}	&	23.4 \D{7.2}	&	17.3 \D{31.5}	&	8.1 \D{67.9}	&	10.0 \D{60.2}	&	21.2 \D{15.7}	&	\Dc{29.4} \\
% imagenet ref: 66.4  &  avg: \Dc{32.8}
\hline
MoCo-b	&	cifar10   	&	83.6	&	76.1 \D{9.0}	&	67.1 \D{19.7}	&	57.2 \D{31.5}	&	47.2 \D{43.5}	&	45.6 \D{45.5}	&	51.0 \D{38.9}	&	\Dc{31.3} \\
MoCo-b	&	cifar100  	&	59.5	&	49.9 \D{16.1}	&	42.4 \D{28.7}	&	32.9 \D{44.7}	&	21.4 \D{64.0}	&	21.1 \D{64.5}	&	27.9 \D{53.0}	&	\Dc{45.2} \\
MoCo-b	&	stl10     	&	95.3	&	92.8 \D{2.6}	&	85.1 \D{10.6}	&	76.0 \D{20.2}	&	38.9 \D{59.2}	&	40.4 \D{57.6}	&	60.9 \D{36.1}	&	\Dc{31.0} \\
MoCo-b	&	cars196   	&	13.8	&	13.7 \D{1.3}	&	12.1 \D{12.8}	&	7.9 \D{42.7}	&	1.8 \D{86.9}	&	1.9 \D{86.3}	&	12.7 \D{8.4}	&	\Dc{39.7} \\
MoCo-b	&	aircraft70	&	26.8	&	28.1 \D{-4.9}	&	26.4 \D{1.5}	&	17.2 \D{35.8}	&	6.9 \D{74.4}	&	8.6 \D{67.8}	&	23.3 \D{13.2}	&	\Dc{31.3} \\
% imagenet ref: 71.1  &  avg: \Dc{35.7}
\hline
SimCLR2	&	cifar10   	&	85.4	&	79.2 \D{7.3}	&	67.5 \D{21.0}	&	58.0 \D{32.1}	&	45.6 \D{46.6}	&	45.8 \D{46.4}	&	54.8 \D{35.8}	&	\Dc{31.5} \\
SimCLR2	&	cifar100  	&	63.5	&	55.2 \D{13.1}	&	44.6 \D{29.8}	&	33.2 \D{47.7}	&	21.2 \D{66.7}	&	22.4 \D{64.7}	&	31.4 \D{50.6}	&	\Dc{45.4} \\
SimCLR2	&	stl10     	&	91.9	&	89.3 \D{2.9}	&	81.7 \D{11.2}	&	69.8 \D{24.1}	&	38.4 \D{58.2}	&	40.8 \D{55.6}	&	61.5 \D{33.1}	&	\Dc{30.8} \\
SimCLR2	&	cars196   	&	17.7	&	16.9 \D{4.6}	&	15.6 \D{11.9}	&	9.8 \D{44.5}	&	1.6 \D{91.2}	&	2.4 \D{86.7}	&	14.3 \D{19.1}	&	\Dc{43.0} \\
SimCLR2	&	aircraft70	&	31.2	&	32.0 \D{-2.4}	&	32.0 \D{-2.3}	&	20.1 \D{35.7}	&	7.1 \D{77.4}	&	10.2 \D{67.2}	&	26.7 \D{14.6}	&	\Dc{31.7} \\
% imagenet ref: 71.0  &  avg: \Dc{36.5}
\hline
BarlowTwins	&	cifar10   	&	83.8	&	77.8 \D{7.1}	&	70.0 \D{16.4}	&	62.0 \D{26.0}	&	51.6 \D{38.5}	&	50.0 \D{40.3}	&	56.9 \D{32.1}	&	\Dc{26.7} \\
BarlowTwins	&	cifar100  	&	63.7	&	56.0 \D{12.1}	&	48.1 \D{24.5}	&	38.8 \D{39.2}	&	26.5 \D{58.5}	&	26.6 \D{58.2}	&	34.2 \D{46.3}	&	\Dc{39.8} \\
BarlowTwins	&	stl10     	&	94.5	&	91.6 \D{3.0}	&	83.7 \D{11.4}	&	74.6 \D{21.1}	&	40.0 \D{57.7}	&	44.8 \D{52.6}	&	63.7 \D{32.6}	&	\Dc{29.7} \\
BarlowTwins	&	cars196   	&	23.4	&	23.6 \D{-1.1}	&	20.7 \D{11.4}	&	11.8 \D{49.6}	&	2.7 \D{88.4}	&	2.4 \D{89.8}	&	18.7 \D{19.9}	&	\Dc{43.0} \\
BarlowTwins	&	aircraft70	&	39.2	&	43.1 \D{-9.7}	&	40.4 \D{-2.9}	&	22.5 \D{42.7}	&	7.7 \D{80.3}	&	9.4 \D{76.1}	&	31.4 \D{19.9}	&	\Dc{34.4} \\
% imagenet ref: 73.5  &  avg: \Dc{34.7}
\hline
DeepCluster	&	cifar10   	&	87.2	&	80.5 \D{7.7}	&	70.6 \D{19.0}	&	64.3 \D{26.2}	&	52.5 \D{39.7}	&	50.3 \D{42.3}	&	57.3 \D{34.3}	&	\Dc{28.2} \\
DeepCluster	&	cifar100  	&	65.0	&	56.2 \D{13.6}	&	47.3 \D{27.1}	&	39.6 \D{39.1}	&	27.7 \D{57.4}	&	25.6 \D{60.6}	&	33.5 \D{48.5}	&	\Dc{41.1} \\
DeepCluster	&	stl10     	&	94.8	&	92.4 \D{2.6}	&	84.6 \D{10.8}	&	79.2 \D{16.5}	&	41.9 \D{55.8}	&	45.0 \D{52.6}	&	64.0 \D{32.5}	&	\Dc{28.5} \\
DeepCluster	&	cars196   	&	22.7	&	20.9 \D{7.8}	&	19.3 \D{15.0}	&	13.8 \D{39.3}	&	2.9 \D{87.2}	&	3.8 \D{83.2}	&	16.7 \D{26.4}	&	\Dc{43.2} \\
DeepCluster	&	aircraft70	&	40.3	&	39.2 \D{2.8}	&	37.6 \D{6.7}	&	24.4 \D{39.5}	&	7.3 \D{81.8}	&	11.2 \D{72.2}	&	27.9 \D{30.7}	&	\Dc{38.9} \\
% imagenet ref: 75.2  &  avg: \Dc{36.0}
\hline
SwAV	&	cifar10   	&	83.5	&	76.8 \D{8.1}	&	68.8 \D{17.7}	&	63.3 \D{24.1}	&	52.9 \D{36.6}	&	48.6 \D{41.8}	&	55.3 \D{33.8}	&	\Dc{27.0} \\
SwAV	&	cifar100  	&	60.1	&	52.5 \D{12.7}	&	44.3 \D{26.3}	&	38.5 \D{35.9}	&	27.1 \D{55.0}	&	23.8 \D{60.4}	&	31.0 \D{48.4}	&	\Dc{39.8} \\
SwAV	&	stl10     	&	94.4	&	91.8 \D{2.8}	&	84.0 \D{11.1}	&	80.3 \D{14.9}	&	43.0 \D{54.5}	&	44.5 \D{52.8}	&	62.7 \D{33.6}	&	\Dc{28.3} \\
SwAV	&	cars196   	&	17.2	&	16.2 \D{5.8}	&	14.6 \D{15.0}	&	12.0 \D{30.4}	&	3.0 \D{82.8}	&	3.0 \D{82.5}	&	12.6 \D{26.9}	&	\Dc{40.6} \\
SwAV	&	aircraft70	&	31.5	&	29.7 \D{5.6}	&	30.5 \D{3.0}	&	23.9 \D{24.2}	&	8.3 \D{73.7}	&	10.3 \D{67.2}	&	22.1 \D{29.7}	&	\Dc{33.9} \\
% imagenet ref: 72.0  &  avg: \Dc{33.9}
\hline
SwAV-b	&	cifar10   	&	84.7	&	78.0 \D{7.9}	&	70.1 \D{17.2}	&	63.4 \D{25.1}	&	52.6 \D{37.8}	&	51.0 \D{39.8}	&	56.7 \D{33.0}	&	\Dc{26.8} \\
SwAV-b	&	cifar100  	&	62.7	&	54.4 \D{13.2}	&	46.5 \D{25.8}	&	39.7 \D{36.6}	&	28.0 \D{55.3}	&	26.5 \D{57.8}	&	33.2 \D{47.1}	&	\Dc{39.3} \\
SwAV-b	&	stl10     	&	94.3	&	91.6 \D{2.9}	&	83.7 \D{11.3}	&	78.5 \D{16.8}	&	44.6 \D{52.8}	&	45.0 \D{52.4}	&	60.9 \D{35.4}	&	\Dc{28.6} \\
SwAV-b	&	cars196   	&	19.3	&	18.0 \D{6.8}	&	16.4 \D{15.2}	&	12.4 \D{36.1}	&	3.1 \D{84.2}	&	3.5 \D{82.0}	&	14.7 \D{24.2}	&	\Dc{41.4} \\
SwAV-b	&	aircraft70	&	33.5	&	32.8 \D{2.1}	&	30.5 \D{9.0}	&	21.7 \D{35.2}	&	8.0 \D{76.0}	&	11.0 \D{67.2}	&	23.9 \D{28.7}	&	\Dc{36.3} \\
% imagenet ref: 74.9  &  avg: \Dc{34.5}
\hline
\end{tabular}
}
\label{tab:downstream_detail}
\end{table*}

\subsection{Downstream robustness test with full fine-tuning}
\label{sec:b.4}

Table~\ref{tab:downstream_resnet} in the main paper and Table~\ref{tab:downstream_detail} above are generated with the KNN evaluation protocol. We also experiment with full fine-tuning on the downstream datasets. The results are in Table~\ref{tab:downstream_finetune}. Since different pre-trained checkpoints are optimized with different optimizers (SGD for Sup, SimSiam\cite{chen2021exploring}, MoCo-v2\cite{chen2020improved}, and SimCLR-v2\cite{chen2020big}; LARS\cite{you2017large} for BYOL\cite{grill2020bootstrap}, BarlowTwins\cite{zbontar2021barlow}, DeepCluster2, and SwAV\cite{caron2020unsupervised}), we use SGD (lr 0.002 cosine) for Sup, SimSiam, MoCo, and SimCLR, and AdamW (lr 0.001 cosine) \cite{loshchilov2018decoupled} for others during fine-tuning. All models are fine-tuned for 10 epochs. We find this strategy of using different optimizers is able to make the baseline results on original images comparable across methods.
We note that fine-tuning drastically improves the accuracy on downstream datasets, while the general observation that CL methods are more robust to downstream corruption than SL still holds, except for BarlowTwins which is slightly worse than SL. Another interesting observation here is that \emph{different CL methods actually yield different robustness behaviors}, although they are all doing some form of contrastive learning and have similar baseline accuracies.

\def\MinNumber{10.0}
\def\MaxNumber{95.0}

\begin{table*}[p]
\caption{Robustness to downstream data corruption with \emph{fine-tuning}. We fine-tune the full network and linear classification layer for 10 epochs. Overall, CL methods are more robust than Sup under this setting except for BarlowTwins.}
\footnotesize
\centering
\setlength{\tabcolsep}{3pt}
\scalebox{0.95}{%
\begin{tabular}{lllllllll c}
\hline
Pre-train Alg & Dataset & Orig & $\gamma=0.2$ &$\gamma=5$ & G-small & G-large & L-small & L-large & Avg $\Delta$ \\
\hline
Sup	&	cifar10   	&	96.7	&	96.8 \D{-0.2}	&	94.0 \D{2.8}	&	88.2 \D{8.8}	&	77.5 \D{19.8}	&	72.0 \D{25.5}	&	86.0 \D{11.0}	&	\Dc{11.3} \\
Sup	&	cifar100  	&	83.8	&	83.7 \D{0.1}	&	77.4 \D{7.6}	&	68.8 \D{17.9}	&	53.5 \D{36.2}	&	46.0 \D{45.1}	&	65.2 \D{22.2}	&	\Dc{21.5} \\
Sup	&	stl10     	&	97.7	&	97.2 \D{0.6}	&	92.5 \D{5.4}	&	92.1 \D{5.7}	&	55.5 \D{43.2}	&	56.5 \D{42.2}	&	89.1 \D{8.8}	&	\Dc{17.6} \\
Sup	&	cars196   	&	75.1	&	73.2 \D{2.6}	&	58.4 \D{22.3}	&	40.1 \D{46.5}	&	4.1 \D{94.6}	&	5.0 \D{93.4}	&	56.5 \D{24.7}	&	\Dc{47.3} \\
Sup	&	aircraft70	&	81.5	&	80.4 \D{1.3}	&	78.8 \D{3.3}	&	66.5 \D{18.4}	&	11.5 \D{85.9}	&	17.6 \D{78.4}	&	72.8 \D{10.6}	&	\Dc{33.0} \\
% imagenet ref: 76.1  &  avg: \Dc{26.2}
\hline
BYOL	&	cifar10   	&	96.5	&	96.3 \D{0.2}	&	93.9 \D{2.7}	&	88.8 \D{7.9}	&	80.2 \D{16.9}	&	75.2 \D{22.0}	&	87.4 \D{9.4}	&	\Dc{9.8} \\
BYOL	&	cifar100  	&	83.2	&	82.2 \D{1.2}	&	76.7 \D{7.8}	&	68.3 \D{17.9}	&	54.2 \D{34.8}	&	46.5 \D{44.1}	&	64.4 \D{22.6}	&	\Dc{21.4} \\
BYOL	&	stl10     	&	96.2	&	95.8 \D{0.5}	&	91.8 \D{4.7}	&	91.3 \D{5.2}	&	57.0 \D{40.8}	&	56.2 \D{41.6}	&	88.2 \D{8.4}	&	\Dc{16.9} \\
BYOL	&	cars196   	&	80.4	&	77.2 \D{4.0}	&	62.1 \D{22.8}	&	49.0 \D{39.1}	&	2.8 \D{96.6}	&	3.9 \D{95.2}	&	65.1 \D{19.0}	&	\Dc{46.1} \\
BYOL	&	aircraft70	&	87.7	&	86.5 \D{1.4}	&	84.1 \D{4.2}	&	76.3 \D{13.0}	&	13.9 \D{84.1}	&	20.2 \D{76.9}	&	80.0 \D{8.8}	&	\Dc{31.4} \\
% imagenet ref: 72.3  &  avg: \Dc{25.1}
\hline
SimSiam	&	cifar10   	&	95.0	&	95.1 \D{-0.1}	&	92.1 \D{3.1}	&	87.5 \D{7.9}	&	79.8 \D{16.1}	&	75.4 \D{20.7}	&	86.7 \D{8.8}	&	\Dc{9.4} \\
SimSiam	&	cifar100  	&	81.0	&	80.5 \D{0.6}	&	74.1 \D{8.5}	&	68.5 \D{15.5}	&	56.4 \D{30.4}	&	51.3 \D{36.6}	&	67.1 \D{17.1}	&	\Dc{18.1} \\
SimSiam	&	stl10     	&	94.0	&	93.4 \D{0.6}	&	88.1 \D{6.3}	&	87.1 \D{7.3}	&	64.4 \D{31.5}	&	59.5 \D{36.7}	&	85.8 \D{8.7}	&	\Dc{15.2} \\
SimSiam	&	cars196   	&	85.7	&	85.2 \D{0.6}	&	75.7 \D{11.6}	&	64.4 \D{24.9}	&	4.2 \D{95.1}	&	5.9 \D{93.1}	&	79.3 \D{7.5}	&	\Dc{38.8} \\
SimSiam	&	aircraft70	&	89.7	&	89.0 \D{0.8}	&	86.9 \D{3.2}	&	82.3 \D{8.3}	&	23.5 \D{73.8}	&	28.9 \D{67.7}	&	86.4 \D{3.7}	&	\Dc{26.3} \\
% imagenet ref: 68.3  &  avg: \Dc{21.6}
\hline
MoCo-b	&	cifar10   	&	96.8	&	96.7 \D{0.2}	&	94.5 \D{2.4}	&	89.6 \D{7.5}	&	81.5 \D{15.9}	&	77.4 \D{20.1}	&	89.4 \D{7.7}	&	\Dc{9.0} \\
MoCo-b	&	cifar100  	&	84.8	&	84.1 \D{0.9}	&	78.5 \D{7.5}	&	72.2 \D{14.8}	&	59.0 \D{30.5}	&	53.9 \D{36.4}	&	70.8 \D{16.5}	&	\Dc{17.8} \\
MoCo-b	&	stl10     	&	96.3	&	96.3 \D{0.0}	&	91.8 \D{4.6}	&	91.6 \D{4.9}	&	64.3 \D{33.2}	&	61.0 \D{36.7}	&	90.3 \D{6.2}	&	\Dc{14.3} \\
MoCo-b	&	cars196   	&	85.7	&	84.6 \D{1.3}	&	75.7 \D{11.7}	&	62.8 \D{26.7}	&	3.6 \D{95.8}	&	5.0 \D{94.2}	&	78.5 \D{8.3}	&	\Dc{39.7} \\
MoCo-b	&	aircraft70	&	90.3	&	89.3 \D{1.1}	&	88.0 \D{2.5}	&	82.5 \D{8.6}	&	22.6 \D{75.0}	&	27.2 \D{69.9}	&	86.9 \D{3.8}	&	\Dc{26.8} \\
% imagenet ref: 71.1  &  avg: \Dc{21.5}
\hline
SimCLR2	&	cifar10   	&	96.3	&	95.8 \D{0.5}	&	93.3 \D{3.1}	&	87.1 \D{9.6}	&	76.8 \D{20.3}	&	72.2 \D{25.0}	&	86.2 \D{10.5}	&	\Dc{11.5} \\
SimCLR2	&	cifar100  	&	84.8	&	84.2 \D{0.7}	&	78.6 \D{7.3}	&	69.5 \D{18.1}	&	56.7 \D{33.2}	&	51.4 \D{39.4}	&	67.3 \D{20.7}	&	\Dc{19.9} \\
SimCLR2	&	stl10     	&	95.5	&	95.2 \D{0.3}	&	89.7 \D{6.0}	&	86.5 \D{9.4}	&	54.6 \D{42.8}	&	55.8 \D{41.6}	&	88.0 \D{7.8}	&	\Dc{18.0} \\
SimCLR2	&	cars196   	&	77.9	&	75.3 \D{3.4}	&	64.9 \D{16.8}	&	47.0 \D{39.6}	&	3.0 \D{96.2}	&	4.5 \D{94.3}	&	68.1 \D{12.6}	&	\Dc{43.8} \\
SimCLR2	&	aircraft70	&	84.8	&	83.8 \D{1.1}	&	82.9 \D{2.2}	&	72.5 \D{14.5}	&	20.1 \D{76.3}	&	23.8 \D{72.0}	&	79.4 \D{6.3}	&	\Dc{28.7} \\
% imagenet ref: 71.0  &  avg: \Dc{24.4}
\hline
BarlowTwins	&	cifar10   	&	96.8	&	96.7 \D{0.1}	&	94.4 \D{2.5}	&	87.9 \D{9.2}	&	76.4 \D{21.0}	&	70.1 \D{27.6}	&	84.9 \D{12.3}	&	\Dc{12.1} \\
BarlowTwins	&	cifar100  	&	83.9	&	83.6 \D{0.4}	&	76.9 \D{8.4}	&	64.2 \D{23.5}	&	46.1 \D{45.1}	&	39.0 \D{53.5}	&	56.4 \D{32.8}	&	\Dc{27.2} \\
BarlowTwins	&	stl10     	&	97.3	&	96.8 \D{0.6}	&	92.2 \D{5.2}	&	91.2 \D{6.3}	&	52.9 \D{45.7}	&	52.0 \D{46.6}	&	87.1 \D{10.5}	&	\Dc{19.1} \\
BarlowTwins	&	cars196   	&	73.5	&	69.0 \D{6.3}	&	53.2 \D{27.7}	&	38.0 \D{48.3}	&	2.7 \D{96.4}	&	3.4 \D{95.3}	&	57.1 \D{22.4}	&	\Dc{49.4} \\
BarlowTwins	&	aircraft70	&	81.1	&	77.9 \D{4.0}	&	76.5 \D{5.7}	&	63.8 \D{21.3}	&	11.3 \D{86.0}	&	15.5 \D{80.9}	&	67.7 \D{16.5}	&	\Dc{35.7} \\
% imagenet ref: 73.5  &  avg: \Dc{28.7}
\hline
DeepCluster2-b	&	cifar10   	&	96.5	&	96.5 \D{0.0}	&	94.6 \D{2.0}	&	89.9 \D{6.9}	&	80.8 \D{16.3}	&	75.7 \D{21.6}	&	87.7 \D{9.2}	&	\Dc{9.3} \\
DeepCluster2-b	&	cifar100  	&	84.7	&	83.6 \D{1.3}	&	78.3 \D{7.5}	&	71.6 \D{15.4}	&	57.6 \D{32.0}	&	49.2 \D{41.9}	&	66.8 \D{21.1}	&	\Dc{19.9} \\
DeepCluster2-b	&	stl10     	&	96.8	&	96.3 \D{0.4}	&	93.7 \D{3.2}	&	93.1 \D{3.8}	&	62.3 \D{35.6}	&	57.6 \D{40.4}	&	88.9 \D{8.1}	&	\Dc{15.3} \\
DeepCluster2-b	&	cars196   	&	81.6	&	79.4 \D{2.6}	&	68.6 \D{16.0}	&	56.3 \D{31.0}	&	3.4 \D{95.8}	&	4.9 \D{94.0}	&	66.5 \D{18.5}	&	\Dc{43.0} \\
DeepCluster2-b	&	aircraft70	&	87.9	&	87.2 \D{0.8}	&	85.4 \D{2.9}	&	77.8 \D{11.5}	&	15.4 \D{82.5}	&	20.0 \D{77.2}	&	79.1 \D{10.1}	&	\Dc{30.8} \\
% imagenet ref: 75.2  &  avg: \Dc{23.7}
\hline
SwAV-b	&	cifar10   	&	96.3	&	96.4 \D{-0.2}	&	94.0 \D{2.3}	&	89.8 \D{6.7}	&	81.6 \D{15.2}	&	75.9 \D{21.2}	&	87.8 \D{8.7}	&	\Dc{9.0} \\
SwAV-b	&	cifar100  	&	83.7	&	83.1 \D{0.7}	&	77.4 \D{7.5}	&	70.8 \D{15.4}	&	58.1 \D{30.5}	&	49.7 \D{40.6}	&	66.3 \D{20.8}	&	\Dc{19.3} \\
SwAV-b	&	stl10     	&	96.3	&	96.6 \D{-0.3}	&	92.8 \D{3.7}	&	92.7 \D{3.8}	&	63.2 \D{34.3}	&	58.9 \D{38.9}	&	88.6 \D{8.0}	&	\Dc{14.7} \\
SwAV-b	&	cars196   	&	82.2	&	80.1 \D{2.5}	&	70.5 \D{14.2}	&	60.4 \D{26.5}	&	3.8 \D{95.4}	&	5.4 \D{93.4}	&	67.7 \D{17.6}	&	\Dc{41.6} \\
SwAV-b	&	aircraft70	&	89.2	&	88.2 \D{1.2}	&	87.2 \D{2.3}	&	80.0 \D{10.4}	&	18.2 \D{79.6}	&	22.9 \D{74.3}	&	81.3 \D{8.9}	&	\Dc{29.5} \\
% imagenet ref: 74.9  &  avg: \Dc{22.8}
\hline
\end{tabular}
}
\label{tab:downstream_finetune}
\end{table*}

\subsection{Variance of pre-training results}

We repeat MoCo-v2 on the original CIFAR-10 200ep three times: The KNN evaluation mean and std is $82.44\pm 0.18$.
Repeating MoCo-v2 on the global 8x8 shuffling corrupted CIFAR-10 gives KNN evaluation mean and std $59.24\pm 0.40$. The linear evaluation variance is similar.
The randomness has a smaller order than the gap between MoCo and Sup results.

\subsection{Pre-train on corrupted CIFAR-10, but test on uncorrupted images}

In the main paper, we show the results when both the pre-training and evaluation datasets are corrupted in the same consistent way. In the following Table~\ref{tab:shuf_uncorrupt}, we report the accuracy numbers obtained from KNN evaluation on the original uncorrupted images. Since these models are pre-trained on the pixel- or patch-level corrupted dataset, the results reflect the transfer capability of the pre-trained representation from corrupted data to original data. We find that the trend is similar to evaluating on corrupted data that Sup appears more robust.

\begin{table}[H]
\caption{Uncorrupted evaluation results of robustness to pre-training pixel-level gamma distortion and patch-level corruption (global and local shuffling) with CIFAR-10 and ResNet-18.}
\centering
\footnotesize
\setlength{\tabcolsep}{2.5pt}
\scalebox{0.9}{%
\begin{tabular}{l|c|ccccc|c}
\hline
\textbf{Method} & Orig  & $\gamma=0.2$ & G4x4 & G8x8 & L4x4 & L8x8 & Avg $\Delta$ \\
\hline
{Sup} & 92.23 & 82.72   & 63.03 & 36.94 & 61.56 & 62.51 & - \\
$\Delta$     & -     & 10.31\% & 31.66\% &	59.95\% &	33.25\% &	32.22\%   & 33.48\%  \\
\hline
{MoCo-v2 KNN} & 82.55 & 72.01   & 46.66 & 32.93 & 48.91 & 53.78 & - \\
$\Delta$           & -     & 15.40\% & 43.48\% &	60.11\% &	40.75\% &	34.85\%   & 38.39\%  \\
\hline
\end{tabular}
}
\label{tab:shuf_uncorrupt}
\end{table}

\section{Additional Visualization}

\subsection{Visualizing corrupted images}

\begin{figure*}[ht]
    \centering
    \includegraphics[width=0.99\linewidth]{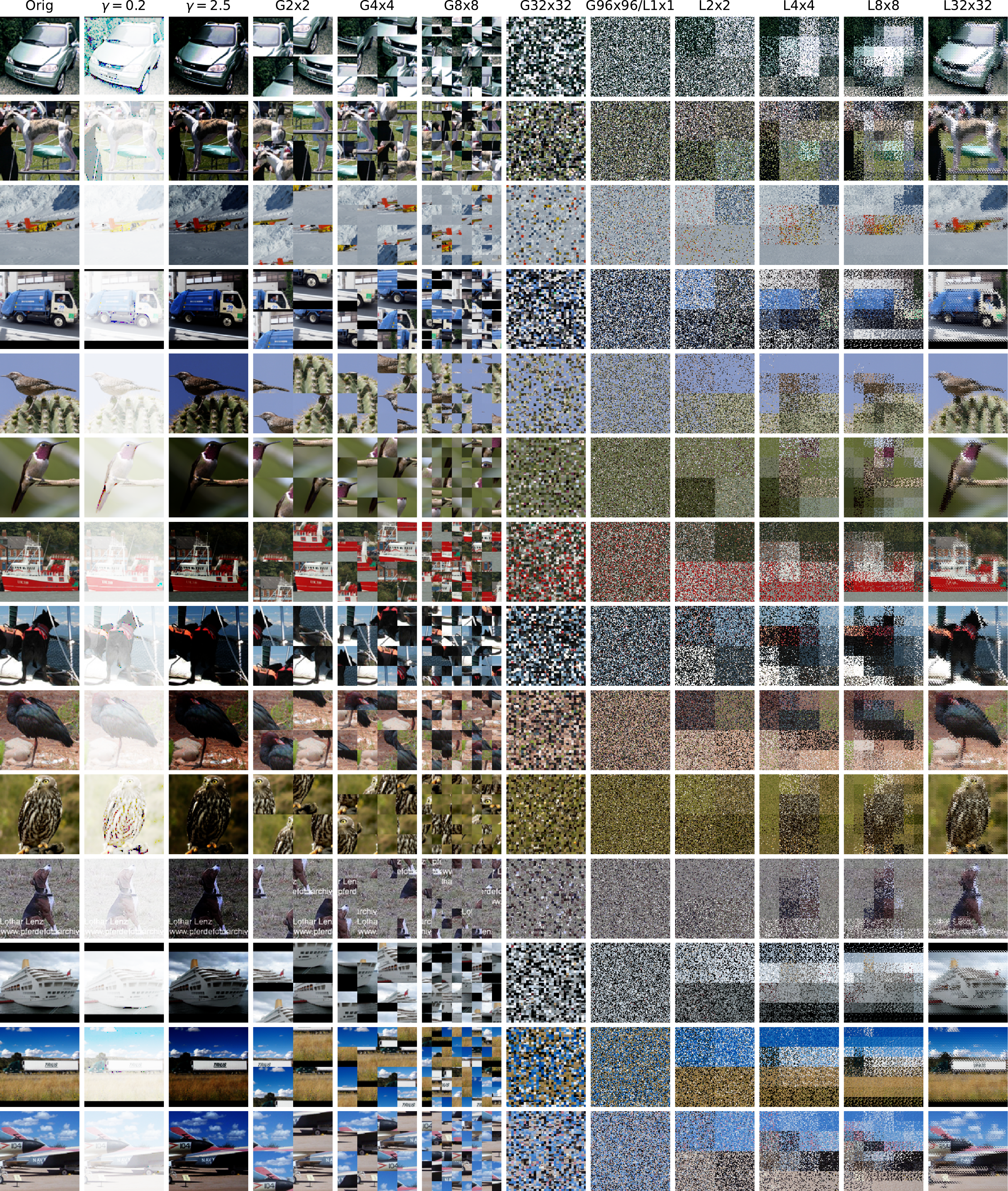}
    \caption{Randomly chosen examples from the STL-10 dataset. The original images have resolution 96x96. We show the resulting images of gamma distortion ($\gamma=0.2,2.5)$, global shuffling (G2x2, weaker -- G96x96, stronger), and local shuffling (L1x1, stronger -- L32x32, weaker). G1x1 and L96x96 revert to the original, while G96x96 and L1x1 are the most random ones (and have similar effect). Gamma distortion reduces information in pixel intensity. Global shuffling destroys global but preserves local structure, while local shuffling is the opposite.}
    \label{fig:vis}
\end{figure*}

Please check Figure~\ref{fig:vis} for more visual examples of the pixel-level gamma distortion and patch-level shuffling corruptions we used.

\subsection{Visualizing Grad-CAM attention maps}

\begin{figure*}[p]
    \centering
    \includegraphics[width=\linewidth]{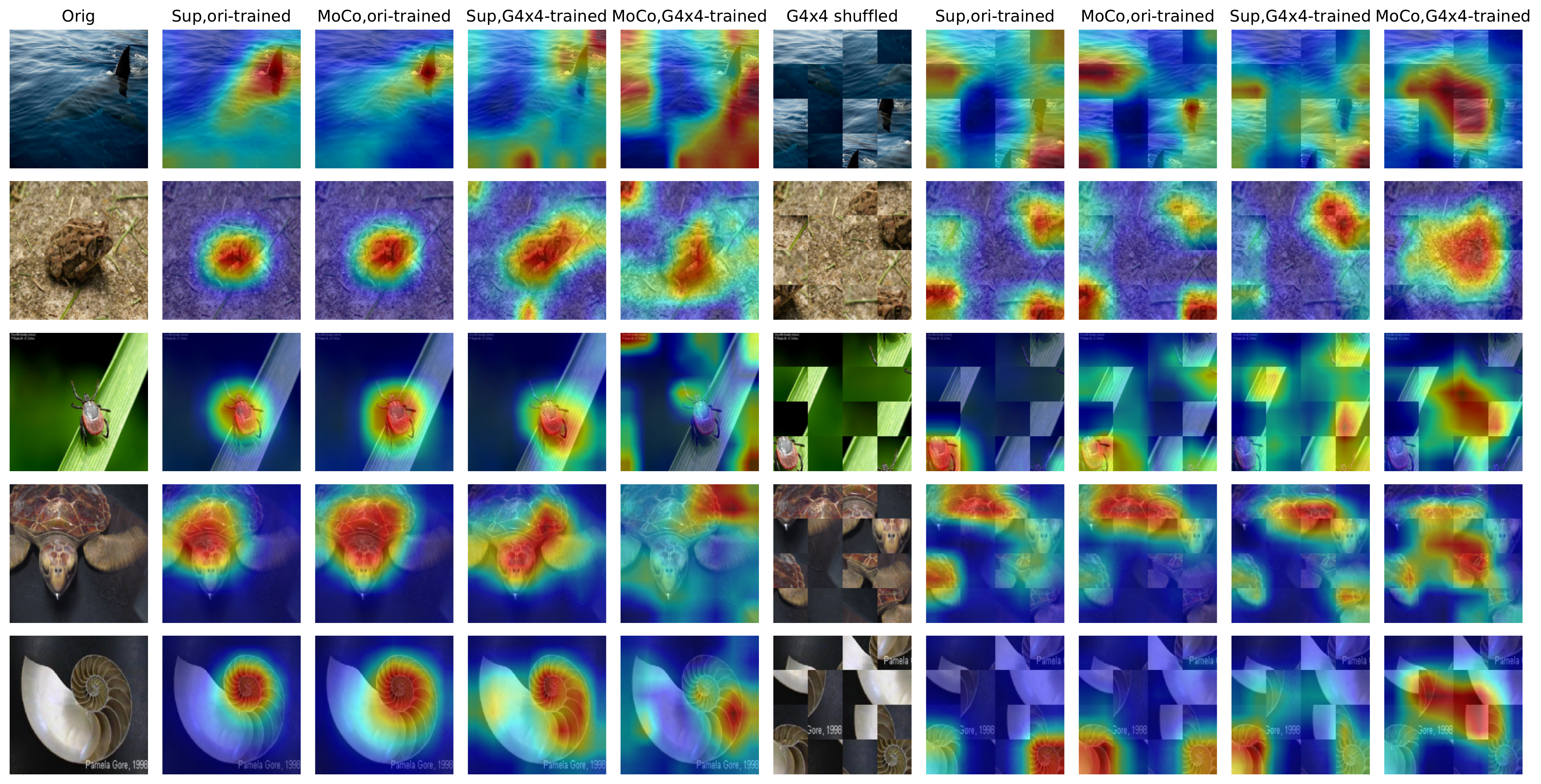}
    \caption{Randomly chosen images from ImageNet-100. We consider 4x4 global patch shuffling and visualize the Grad-CAM attention maps of 4 models: Sup trained on original images, MoCo trained on original images, Sup trained on shuffled images, and MoCo trained on shuffled images. The attention map of the MoCo model on shuffled images is less equivariant to the patch shuffling.}
    \label{fig:gradcam}
\end{figure*}

Figure~\ref{fig:gradcam} visualizes the Grad-CAM \cite{selvaraju2017grad} attention maps of ResNet-18 models pre-trained and linearly fine-tuned on either uncorrupted or 4x4 global patch shuffled images. We discover some difference in terms of the \emph{equivariant} property: Sup models are largely equivariant to 4x4 global patch shuffling -- the attention is focused on the object parts even after patch shuffling, whereas the MoCo model pre-trained on 4x4 global shuffled images are not -- it is rather focused on distracting parts. The quality of attention maps correlates with the top-1 validation accuracy, where Sup on 4x4 achieves 65\% and MoCo achieves 35\%. Intuitively, a model can be more robust to the global patch shuffling if it possesses such an equivariant property. This shows the robustness of SL from another aspect, because it can robustly learn the same feature even under the shuffling disturbance.

\end{document}